\documentclass[twoside,11pt]{article}
\usepackage{jmlr2e}
\usepackage{amsmath,amsbsy}
\usepackage{algorithm,algorithmic,caption}
\usepackage{mathtools}
\newcommand\defeq{\coloneqq}
\newcommand{\E}{{\mathbb E}}
\newcommand{\V}{{\mathbb V}}

\ShortHeadings{Variational methods and linear response}{Raymond and Ricci-Tersenghi}
\firstpageno{1}

\begin{document}

\title{Improving variational methods via pairwise linear response identities}

\author{\name Jack Raymond \email jack.raymond@physics.org \\
       \addr Dipartimento di Fisica, La Sapienza University of Rome\\
       Piazzale Aldo Moro 5\\
       Rome, Italy
       \AND
       \name Federico Ricci-Tersenghi \email federico.ricci@uniroma1.it \\
       \addr Dipartimento di Fisica, INFN--Sezione di Roma1 and CNR--Nanotec,\\
       La Sapienza University of Rome\\
       Piazzale Aldo Moro 5\\
       Rome, Italy
       }

\editor{}

\maketitle

\begin{abstract}%   <- trailing '%' for backward compatibility of .sty file
Inference methods are often formulated as variational approximations: these approxima-
tions allow easy evaluation of statistics by marginalization or linear response, but these
estimates can be inconsistent. We show that by introducing constraints on covariance, one
can ensure consistency of linear response with the variational parameters, and in so doing
inference of marginal probability distributions is improved. For the Bethe approximation
and its generalizations, improvements are achieved with simple choices of the constraints.
The approximations are presented as variational frameworks; iterative procedures
related to message passing are provided for finding the minima.
\end{abstract}


\section{Introduction}

%Inference methods are often formulated as variational approximations: these approximations allow easy evaluation of statistics by %marginalization or linear response, but these
%estimates can be inconsistent. We show that by introducing constraints on covariance, one
%can ensure consistency of linear response with the variational parameters, and in so doing
%inference of marginal probability distributions is improved. For the Bethe approximation
%and its generalizations, improvements are achieved with simple choices of the constraints.
%The approximations are presented as variational frameworks; iterative procedures
%related to message passing are provided for finding the minima.

Given a probability distribution $p(x)$, estimation of marginal probability distributions such as $p(x_i)$
and $p(x_i; x_j)$ is one of the most important inference tasks addressed in graphical models, alongside estimation of the
maximum probability state and the log partition function~[\cite{Wainwright:GME,Mezard:IPC,MacKay:ITI}]. The challenge is addressed in many research
fields by a variety of methods. In Boltzmann machine learning and probabilistic independent component analysis the
expectation-maximization algorithm requires such estimates~[\cite{Wainwright:GME,Miskin:AIC}]. In heuristic optimization, a branch and bound search (or decimation procedure) over a high dimensional space can be made more efficient, by branching on $x_i$ (or some small set of variables) in an informed manner using approximate probabilities~[\cite{Montanari:SCS}]. In channel coding we wish to determine the likely state of a bit sent over a noisy channel, which can be inferred with a measure of certainty from the marginal probability~[\cite{Richardson:MCT}]. In statistical physics, marginal probability distributions provide insight into phase transitions and thermodynamic phases~[\cite{Parisi:SFT}].

Approximation of marginal probability distributions (called marginals henceforth) with
high accuracy is NP-hard even in the case of Ising spins (binary variables) with pairwise
interactions~[\cite{Dagum:ApI,Long:RBM}], but in practice, many schemes might be applied
successfully. Approximate inference of marginal distributions is often performed by Markov
chain Monte Carlo (MCMC) procedures~[\cite{Andrieu2003}]. These methods are a workhorse of inference,
but have some disadvantages: the estimates are achieved with an accuracy that decays only
slowly with time resources (exploiting the central limit theorem), the result is stochastic, and
takes a non-parametric form. For these reasons variational approximations are often preferred,
the price being a (difficult to quantify) bias in the approximations~[\cite{Wainwright:GME,Mezard:IPC,MacKay:ITI}]. 

In a basic variational approximation an intractable probability distribution $p(x)$ is approximated by a tractable one $q(x)$, the parameters of $q$ are determined by minimizing
the Kullback Leibler (KL) divergence $D_{KL}[p|| q]$. The challenge of marginalization is thus
replaced by two linked challenges: appropriate construction of $q$, and minimization of a
KL-divergence. An example of a tractable distribution is a factorized one: $q(x)=\prod_i q(x_i)$,
which leads to a mean-field variational approximation; KL-divergence could then typically
be minimized by an iteration of fixed point equations~[\cite{Wainwright:GME,Mezard:IPC}]. 
It is common for the estimates obtained by variational
approximations to be over-confident, the uncertainty in some variables is reduced since the
structure of the approximation $q$ discounts some sources of variance.

A given variational framework may be minimized by several algorithms, and it is interesting that many famous heuristic algorithms developed independently of variational
frameworks have been shown to be particular solutions to variational approximations. Most
notably loopy belief propagation has been shown to be one method to solve the Bethe
variational approximation, and expectation propagation was shown to be one method to
solve the expectation consistent variational approximation. This connection to variational
frameworks has allowed interesting insight into algorithm construction, proofs of solution
existence and convergence~[\cite{yedidia,Wainwright:GME,Yuille:cccpalgorithms}].

We have so far identified one mechanism by which to estimate marginals: we can directly marginalize our approximation $q(x)$, which is tractable by construction. These same
estimates (when not extremal) can be found as first derivatives of the variational function
(in the above description, the KL-divergence), as we later show. Second order derivatives
of the KL-divergence are also possible and can give covariance estimates. These covariance
estimates are called linear responses - since they measure the response of some expectation
(the first derivative estimate) to an infinitesimal linear perturbation (the second derivative).
Linear response and marginalization estimates are tractable for variational approximations;
and for some statistics, we can use either method to obtain an estimate for the same quantity.

We will consider variational approximations applied to a model of $N$ discrete variables $x_i$ defined by probability~\footnote{For brevity in expressions we will use the notation $x$ both for a random variable and its realization,
relying on context for the distinction.}
\begin{equation}
  p(x) = \frac{1}{Z} \prod_{a=1}^M \psi_a(x_a)\;, \label{eq:probNoh}
\end{equation}
where $\psi_a$ are the potentials (also called factors) and are non-negative functions of the variables indexed by subset $a$, $x_a=\{x_i:i\in a\}$, $Z$ is the partition function. 
Probabilities of this kind can be represented as a factor graph~[\cite{Wainwright:GME,Mezard:IPC}], as shown in Figure \ref{fig:alarmnet}.
Our aim is to demonstrate a mechanism whereby existing variational schemes can be leveraged for improved inference of marginals.

This paper considers a new self-consistent approximation to improve variational methods, with an emphasis on the Bethe approximation and its generalizations (called region-based, or cluster-variational, approximations).
We propose the addition of constraints requiring the consistency of estimates obtained via direct marginalization and linear response. 
We minimize the variational function subject to an agreement of these estimates and show that the resulting unique estimate is an improvement on the two estimates that are obtained without the constraints.

\subsection{Literature review}
The linear response has been leveraged to improve estimates of marginals in a variety of problems, the idea originating in statistical physics~[\cite{Parisi:SFT,Opper:VLR,Welling:LRA}].

Physics approaches often aim to improve understanding of phase transitions for problem
classes in the limit of a large number of variables~[\cite{Parisi:SFT}], rather than in development
of algorithmic approaches to solve particular finite instances. An early application of linear response was the self-consistent Ornstein-Zernike approximation (SCOZA)~[\cite{HoyeSell1977}], which was later applied to simple graphical models [\cite{dickman1996self}]. The SCOZA has been applied to disordered models where marginals are not homogeneous, e.g.\ to the random field Ising model [\cite{Kierlik:SCOZ}], but in the service of estimating globally averaged and disorder averages quantities, and never in such a way as to understand particular single variable or pairwise marginals, which is the question we address.

In 2001 Opper and Winther proposed the adaptive-TAP approach as an extension of the standard mean-field method~[\cite{Opper:TAPMprl}]; in the original formulation of this method, a self-consistency relation between the linear response and magnetizations of a mean-field approximation were reconciled to arrive at a more advanced mean-field theory. This method was later reinterpreted as a special case of expectation consistent approximate inference, that made a connection between moment-matching algorithms such as expectation propagation and variational frameworks, as well as expanding the range of applications ~[\cite{Opper:ECFE,Winther:ECAI}]. Expansions of the expectation consistent approximation to mitigate for errors on higher order cumulants have shown promise, but at this point come with few theoretical guarantees~[\cite{Opper:PC,Paquet:PC}].

In the context of machine learning, other related approaches for improving mean-field estimation have been successfully demonstrated~[\cite{Kappen:BML,Giordano:LRM}]. Mean-field variational Bayes is an important application of variational approximations, but the absence of an accurate understanding of covariance in the model parameters had been a weakness. Recently it was shown how linear response could be used to more accurately estimate these quantities~[\cite{Giordano:LRM}]. 

The Bethe variational approximation is also an important approximation in the context of sparse graphical models, for which loopy belief propagation (LBP) is the most famous algorithm. The linear response has also been used to improve this approximation~[\cite{Montanari:CLC,Mooij:loopcorrected2}]. A large part of this development has been through loop-correction algorithms since the failure of the approximation is known to be related to loops in the graphical model representation. There also exist elegant loop correction methods not relying on the linear response: libDAI is a code repository that has collected some of the methods together~[\cite{Mooij_libDAI_10}], we developed our methods based on this library, in particular, the implementation of~[\cite{Heskes:AICO}]. An important class of loop correction approximations not included in that library are based on the expansion of the log partition function error in loop-terms, loop calculus~[\cite{Chertkov:LC}].

In 2013 several papers related to an extension of the Bethe approximation were published by Yasuda et al and the authors~[\cite{Yasuda:SPDC,Raymond:MFM,Raymond:Correcting,Yasuda:GISP}]. The idea was very similar to that of adaptive-TAP, to minimize the variational function subject to the constraint of statistical consistency. When applied to the mean-field approximation it was realized these methods were equivalent to adaptive-TAP, but in the context of Bethe and region-based free energies improved performance was identified. All these frameworks were also leveraged in the reverse direction to solve the inverse-Ising problem (inferring parameters from statistics)~[\cite{Raymond:MFM,Huang:adaTAP}]. 

These variational frameworks were applied initially to Bethe and mean-field approximations on pairwise binary state models, an extension to general discrete states was provided in~[\cite{Yasuda:GISP}], whereas region-based variational frameworks and a broader range of constraint types were considered in~[\cite{Raymond:MFM}]. In this paper, we consider generic discrete alphabets, region-based free energies (inclusive of the Bethe approximation) and both single-variable and pairwise variable consistency constraints. Building on a belief propagation approach we derive tools for solving the problem with the additional constraints.

\subsection{Outline}
In Section \ref{sec:freeenergies} we define a set of variational approximations; and motivate the inclusion of additional constraints.
In Section \ref{sec:freeenergyminimization} we describe methods to minimize the free energy subject to these constraints.
In Section \ref{sec:results} we compare the performance of constrained approximations against exact results on some standard models; demonstrating a significant advantage in many cases.
In Section \ref{sec:discussion} we discuss our findings in the context of all experimental results and other insights gained, before concluding in Section \ref{sec:conclusion}.
Appendices include exact expressions for the fully connected ferromagnet example, pseudocode and algorithmic details, proofs of convergence for some methods, discussion of solution existence and convergence, and how to select constraints for inclusion.

% Acknowledgments should go at the end, before appendices and references

\section{Constrained variational approximations}
\label{sec:freeenergies}

\begin{figure}
%~/GIT/libdai/utils/fg2dot $file $file.dot
%dot -Tps -Nlabel="" $file.dot -o Aug4$file.ps
\centering
  \includegraphics[width=\linewidth]{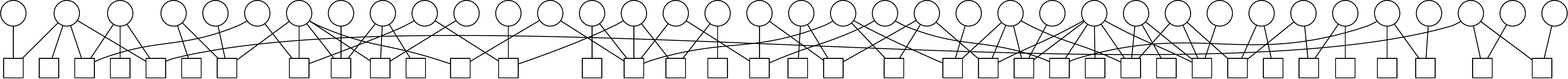}\\
  \vspace{4mm}
  \includegraphics[width=\linewidth]{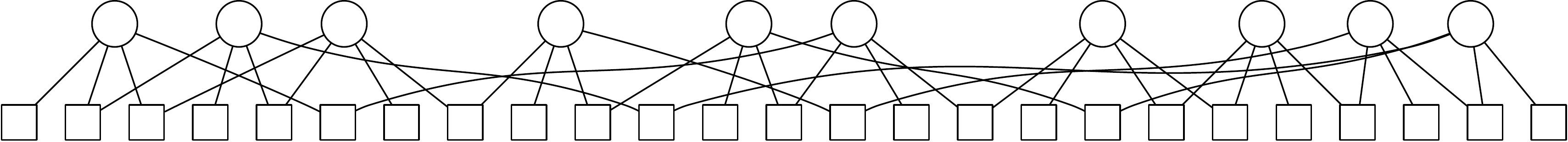}
  \caption{\label{fig:alarmnet} The constrained variational approximations we present can be applied to models with multi-variable interactions, as represented by factor graphs. Two examples are shown. Top: the alarm network is a well-known toy example of a Bayesian net, here represented as a factor graph. Squares denote factors $\psi_a(x_a)$, which act over subsets of variables $x_a$, each variable represented by a circle. Bottom: $N=10$ variables interacting according to a random cubic graph is represented, each coupling ($J$) is represented by a factor with two connections, each field ($h$) by a singly connected factor.}
\end{figure}

Variational free energy approximations are powerful tools for approximate inference~[\cite{yedidia,Opper:ECFE,Wainwright:GME,Mezard:IPC}]. We introduce in this section the mean-field, Bethe, and region-based (also called Kikuchi) approximations. A set of simplified expressions appropriate to the Bethe approximation for an Ising model is given alongside the general expressions. We first introduce a generalization of the probability over the $N$ variables $x$, introducing auxiliary parameters $\nu$,
\begin{equation}
  p_\nu(x) =  \frac{1}{Z(\nu)} \prod_{a=1}^M \psi_a(x_a) \prod_{i=1}^N\left[\prod_y \exp\left( \nu_{i,y} \delta_{x_i,y} \right) \right] \;. \label{eq:prob} 
\end{equation}
This coincides with (\ref{eq:probNoh})  in the limit $\nu\to 0$. The product on $y$ is over all possible states of $x_i$. For our method to apply the probability needs to be differentiable with respect to the parameters $\nu$, modifications would allow for the more general case\footnote{Special care should be taken in cases where $\psi_a(x_a)=0$ for some $x_a$, in such cases it may not be meaningful to perturb by $\nu$. Furthermore, we note that $\nu$ are a redundant set of parameters since variation of $\sum_y \nu_{i,y}$ leaves the probability unchanged. }. The choice of statistics $\{\delta_{x_i,y}\}$ simplifies our presentation, but more generally we might consider a set of single variable functions $\{\phi_y(x_i)\}$, this is discussed in detail in Appendix \ref{app:constraintSelection}.

A Boltzmann machine will be presented as a running example. The model\footnote{For binary variables, we will use Ising spins $s=\pm 1$ in place of binary states $b\in\{0,1\}$. The transformation $s=1-2b$ allows conversion between these two conventions.} has Ising spin variables $x \in \{-1,1\}^N$, fields $h$ and pairwise couplings $J$. A connected graphical model will be assumed for notational simplicity, so each variable has connectivity $k_i \ge 1$, and a unique connected component exists. If an edge set $E=\{(i,j)\}$ specifies the interacting variables
\begin{equation}
  p_\nu(x) = \frac{1}{Z(\nu)} \prod_{(i,j) \in E}\exp(J_{ij}x_i x_j) \prod_i \exp[(h_i+\nu_i) x_i]\;, \label{eq:pising}
\end{equation}
where we use a non-redundant set of auxiliary parameters $\nu_i=\nu_{i,1}-\nu_{i,-1}$.

In a variational approximation a trial probability distribution $q(x)$ is related to the log-partition function of the full model, derived from the Kullback-Leibler divergence 
\begin{equation}
  D_{KL}[q || p_\nu] = \sum_x q(x) \left[\log q(x) - \sum_{i,y} \nu_{i,y}\delta_{x_i,y} - \sum_a \log \psi_a(x_a)\right]  + \log Z(\nu)  \label{KLdivergence1}\;.
\end{equation}
The variational free energy (VFE) is defined
\begin{equation}
  F_\nu(q) = D_{KL}[q || p_\nu] - \log Z(\nu)  \label{KLdivergence} \;,
\end{equation}
which is the tractable part of (\ref{KLdivergence1}).
The optimal variational parameters $q^*$ are those minimizing (\ref{KLdivergence}).

In the simplest mean-field approximation a factorized variational form is considered $q(x) = \prod_{i=1}^N q_i(x_i)$.
The parameters $\{q_i\}$ are precisely marginal distributions on single variables. Iterative methods are often successful in minimizing the VFE.

In another class of approximations the entropy term in the VFE,
$-\sum_x q(x) \log q(x)$, is decomposed as a truncated sum of marginal entropies. In the Bethe
approximation a redundant set of marginal probability distributions $\{q_a(x_a); q_i(x_i)\}$\footnote{$q_A(x_A)$ denotes a variational parameter, that can be interpreted as an approximation to the marginal probability $p(x_A)$.} are
introduced in one-to-one correspondence with the model factors and variables, and the
entropy is approximated as
\begin{equation}
  -\sum_x q(x) \log q(x) \approx -\sum_a \sum_{x_a} q_a(x_a) \log q_a(x_a) - \sum_i (1-k_i) \sum_{x_i}q_i(x_i) \log q_i(x_i) \label{eq:bethefreeenergy}\;,
\end{equation}
with $k_i$ equal to the variable connectivity (the number of factors in which variable $i$ participates). The reason this approximation may improve upon mean-field is that through $q_a(x_a)$
some correlations amongst variables may be explicitely represented, which are absent in the
factorial form of mean-field. The variational parameters, $q_a$ and $q_i$, are referred to as beliefs.

\begin{figure}
\centering
  \includegraphics[width=\linewidth]{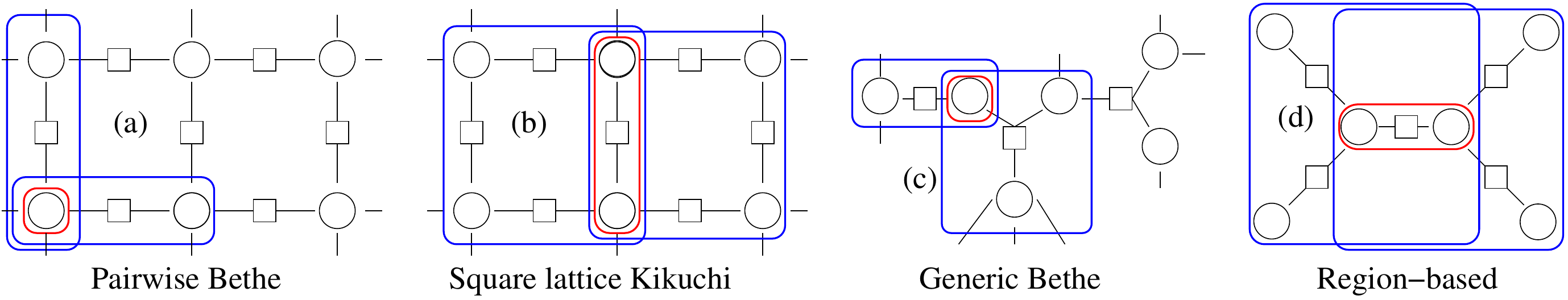}
  \caption{\label{fig:regions} In a region-based approximation (Bethe is a special case), the assignment of outer regions determines the entropy approximation. Sensible choices collect nearby sets of strongly dependent variables. A region graph has outer regions (blue) and intersection regions (red), such that every factor is associated to exactly one outer region; the set of factors in a region $\alpha$ define an auxiliary interaction (\ref{eq:auxiliaryregioninteraction}). Selection (a) is a Bethe approximation, the outer regions are pairs of variables, that intersect on single variables. (b) Alternatively, for a square lattice, we might consider outer regions of 4 variables (the central interaction could be assigned to either the right or left region), that intersect on pairs of variables, which in turn intersect on single variables. This approximation is powerful for square lattice models~[\cite{Raymond:MFM,dominguez2011characterizing,lage2013replica}]. (c) The Bethe approximation on a factor graph with mixed (multi-variable) interaction types has outer regions containing exactly one factor and its variables, the intersections are single variables. (d) On a locally tree-like graph, we can also make an interesting approximation: outer regions are stars that intersect on edge regions. This approximation relates to an alternative loop-correction algorithm~[\cite{Montanari:CLC}].}
\end{figure}

The Bethe approximation is a special case of the more general region-based approximation~[\cite{yedidia}], where the entropy approximation is implied by a choice of outer
regions. Each outer region is defined by a set of variables $x_\alpha$, and a generalized factor $\psi_\alpha(x_\alpha)$ that
describes the interactions between those variables. The outer regions must be chosen to span all
variables, and the factors (as well as auxiliary parameters $\nu$) can be distributed amongst
the generalized factors such that  $\prod_\alpha \psi_\alpha(x) \propto  p_\nu(x)$. Some examples of region selections discussed in
this paper are shown in Figure \ref{fig:regions}.

The entropy approximation is implied by the choice of outer regions: it is the sum of
the entropy on the outer regions $\alpha$ corrected by a weighted sum of entropies on region
intersections ($\beta$). The region based free energy is defined
\begin{equation}
  F_\nu(q) = \sum_\alpha \sum_{x_\alpha} q_\alpha(x_\alpha) \log \left(\frac{q_\alpha(x_\alpha)}{\psi_\alpha(x_\alpha)}\right) + \sum_\beta c_\beta \sum_{x_\beta}q_\beta(x_\beta) \log q_\beta(x_\beta) \label{eq:regionfreeenergy}\;,
\end{equation}
where $c_\beta$ take integer values according to a simple rule~[\cite{yedidia,Heskes:AICO}]. Larger regions are capable of capturing more correlations between variables explicitely, but at a computational cost that scales (in the absence of further approximations) exponentially with region size. This trade-off determines the choice of regions.

In the Bethe approximation, the outer regions $(\alpha)$ are in one-to-one correspondence with the factors ($a$) of the model, and intersection regions are single variables. In our running example of the Boltzmann machine (\ref{eq:pising}), a Bethe approximation has edges as regions. Generalized factors can be chosen as
\begin{equation}
  \psi_{(i,j)}(x_i,x_j) = \exp\left(J_{ij} x_i x_j + \frac{h_i + \nu_i}{k_i} + \frac{h_j + \nu_j}{k_j}\right)\label{eq:auxiliaryregioninteraction}\;.
\end{equation}

We can define marginal variational parameters for each marginal probability in the free energy and then minimize (\ref{eq:regionfreeenergy}) subject to local consistency constraints
\begin{eqnarray}
  \sum_{x_\alpha} q_\alpha(x_\alpha) &=& 1 \;; \qquad \forall \alpha \label{eq:consNorm}\\
  \sum_{x_\alpha\setminus x_\beta} q_\alpha(x_\alpha) &=&  q_\beta(x_\beta) \;; \qquad \forall \alpha, \forall \beta \subset \alpha \label{eq:consMarg}
\end{eqnarray}
Minimization of this free energy is a well-studied problem. Heuristic approaches normally lead to message passing algorithms, which are often convergent to good solutions even where guarantees of convergence are lacking. It is always possible to find minima of the region based free energy using a convex-concave procedure, which is guaranteed to converge to a local minima~[\cite{Yuille:cccpalgorithms,Heskes:AICO}].

Having found a minimum of the free energy at $q=q^*$, we can define the linear response in the beliefs to a perturbation in $\nu_{i,y}$, as
\begin{equation}
  q^*_{\alpha,(i,y)}(x_\alpha) =  \left. \frac{\partial q_\alpha(x_\alpha)}{\partial \nu_{i,y}}\right|_{q=q^*}
\end{equation}
Whereas the variational parameters $q$ have an interpretation away from the fixed point, the linear response is only defined about a global (or heuristically, local) minima. Notation $^*$ will be used to denote an evaluation at such a fixed point.

The entropy approximations (\ref{eq:bethefreeenergy}-\ref{eq:regionfreeenergy}) can be interpreted as truncated series~[\cite{pelizzola:CVM,Wainwright:GME}], that can be made good either by considering sufficiently large regions (those defining a junction tree~[\cite{Wainwright:GME}]), or including loop corrections~[\cite{Chertkov:LC}]. The made-good approaches are not tractable in many interesting models of modest scale. The region based methods most often lead to improvements over mean-field, but entropy expansion can also lead to counterintuitive features. For example, the entropy estimate can be negative, under such an approximation.

\subsection{Inconsistency of covariance approximations}
The covariance of a pair of statistics $\phi_1$ and $\phi_2$, under probability distribution $p$ is defined as
\begin{eqnarray}
  \V_p(\phi_1,\phi_2) &=& \E_p(\phi_1 \phi_2) - \E_p(\phi_1) \E_p(\phi_2)\;,\\
    \text{with}\quad \E_p(\phi) &=& \sum_x p(x) \phi(x)\;.
\end{eqnarray}
For the case $\phi_1(x_\alpha)=\delta_{x_{i_1},y_1}$ and $\phi_2(x_\alpha)=\delta_{x_{i_2},y_2}$, we can replace the probability of interest $p$ by $q_\alpha$ to obtain
\begin{equation} 
\V_p(\delta_{x_{i_1},y_1},\delta_{x_{i_2},y_2}) \approx C_{(i_1,y_1),(i_2,y_2)} \defeq  \V_{q_{\alpha}} [\delta_{x_{i_1},y_1},\delta_{x_{i_2},y_2}]\;. \label{eq:Vp}
\end{equation}
$C$ will be called the marginal approximation to the covariance, and is a function of the variational parameter $q_\alpha$. The optimal value $C^*=C(q^*)$ is independent of the region $\alpha$, owing to the consistency constraints (\ref{eq:consNorm})-(\ref{eq:consMarg}).

Alternatively, we can begin with a second derivative identity. By introducing parameters $\nu$ conjugate to each statistics (\ref{eq:prob}), we have
\begin{equation}
  \V_p(\delta_{x_{i_1},y_1},\delta_{x_{i_2},y_2}) = \frac{\partial^2 \log Z(\nu)}{\partial \nu_{i_1,y_1} \partial \nu_{i_2,y_2}}\;. \label{eq:Vq}
\end{equation}
We obtain a tractable approximation replacing $\log Z(\nu)$ by $-F_\nu(q^*)$
\begin{equation}
  \V_p(\delta_{x_{i_1},y_1},\delta_{x_{i_2},y_2}) \approx \chi_{(i_1,y_1),(i_2,y_2)} \defeq \sum_{x_\alpha} q^*_{\alpha,(i_1,y_1)}(x_\alpha) \delta_{x_{i_2},y_2}\label{eq:Vqstar}\;,
\end{equation}
for any $\alpha$ containing $i_2$, which is called the linear response estimate. $\chi$ will be called the linear response approximation to the covariance, it is defined only at the minima $q^*$, and is a symmetric matrix. We do not need to make explicit reference to the region $\alpha$ used in Eq.~(\ref{eq:Vqstar}) since the linear response $\chi$ does not depend on that choice.

The name `linear response' for the quantity $\chi$ comes from the fact it can be interpreted as
\begin{equation}
\chi_{(i_1,y_1),(i_2,y_2)} \approx
\left.\frac{\partial \E_p[\delta_{x_{i_1},y_1}]}{\partial \nu_{i_2,y_2}}\right|_{\nu=0} =
\left.\frac{\partial \E_p[\delta_{x_{i_2},y_2}]}{\partial \nu_{i_1,y_1}}\right|_{\nu=0}\;,
\end{equation}
that is the linear variation of the mean value of a single variabile statistics to a small perturbation in the the parameter $\nu$ conjugated to another single variable statistics.

We denote the difference of these estimates for two statistics on variables $(i_1,i_2)$ contained in region $\alpha$ %\footnote{quantities for different $\alpha$ are equal when the free energy is minimized.}
 as
\begin{equation}
  \Delta_{(i_1,y_1),(i_2,y_2),\alpha}(q_\alpha,q^*_{\alpha,(i_1,y_1)}) = C_{(i_1,y_1),(i_2,y_2)} - \chi_{(i_1,y_1),(i_2,y_2)} \label{eq:Delta}\;.
\end{equation}
Except for some simple models $C^*-\chi$ is non-zero, exposing an inconsistency in the variational method. The best marginal approximation does not match the best linear response estimate. To decide which estimate to use, either connected correlations or linear responses, we might consider the distance of these two different estimates from the correct value:
\begin{eqnarray}
  \Delta^{(1)}_{(i_1,y_1),(i_2,y_2)} &=& C^*_{(i_1,y_1),(i_2,y_2)} - \V_p\left(\delta_{x_{i_1},y_1},\delta_{x_{i_2},y_2}\right) \label{eq:belieferror}\\
  \Delta^{(2)}_{(i_1,y_1),(i_2,y_2)} &=& \chi_{(i_1,y_1),(i_2,y_2)} - \V_p\left(\delta_{x_{i_1},y_1},\delta_{x_{i_2},y_2}\right) \label{eq:linearresponseerror}
\end{eqnarray}
Graphical models in which the Bethe approximation is most successful have relatively weak correlations and/or few short loops. In these cases it is known that the linear response estimate (\ref{eq:linearresponseerror}) improves significantly upon (\ref{eq:belieferror}) for $i_1 \neq i_2$~[\cite{Welling:LRA,Raymond:Correcting}]. However, as the approximation breaks down (due to poor approximations of loops in the graphical model), the response estimate can be much worse; even giving infinite values for bounded statistics.
For pairs with $i_1=i_2$ (called diagonal), bounds can be violated even in regimes where the approximation is good. A simple example is the model of Section \ref{sec:FC} with zero field ($h=0$): whilst it is true that $\V_p(x_i,x_i) \leq 1$ for any model, $\chi_{i,i}>1$ for in the weakly coupled (high temperature) regime. 

\subsection{Covariance constraints}

We would like to use the linear response information, in a safe manner to select the best
covariance estimate, but also to make the approximation self-consistent. Rather than
simply minimizing the free energy to determine $q^*$, we do this in the subspace where
$\{\Delta_{(i_1,y_1),(i_2,y_2),\alpha} = 0\}$ for a subset of the covariances.

An important question will be which covariances to constrain. Expansion methods indicate that adding all constraints is best when the approximation is very good~[\cite{Raymond:Correcting}]. More generally we wish to add important constraints in so far as it does not prevent solution existence and allows algorithmic stability as discussed in Appendices \ref{app:uniqueness}-\ref{app:constraintAssignment}.

Although general expressions are derived, experimental sections in this paper are restricted principally to the Bethe approximation, and with simple patterns of constraints discussed in Appendix \ref{app:constraintSelection}. The Bethe approximation with the addition of all possible constraints (called {\em on and off diagonal}\footnote{This terminology arises by considering which elements of the covariance matrix are constrained.}) is considered, as well as the Bethe approximation
with addition of only constraints for which $i_1 =  i_2$ ({\em diagonal}). In section 4.1 we also
present results for the mean-field approximation with all possible constraints (since $i_1 = i_2$ in
all cases, this is also called diagonal), as well as the Bethe approximation including only
constraints for which $i_1 \neq i_2$ ({\em off-diagonal}). In other experiments, we do not present these latter
two regimes since they performed consistently worse except in some narrow parameter ranges where all regimes were performing poorly.

\subsection{Bethe approximation to the Boltzmann machine}
In the case of our running example of the Boltzmann machine (\ref{eq:pising}) we are considering variation of $\nu_i=\nu_{i,1}-\nu_{i,-1}$, 
accordingly we can abbreviate notation everywhere $(i,y)$ to $i$. 
In the diagonal constraint approximations we will require consistency of $\V(x_{i},x_{i})\;\forall \;i$, 
in the on and off diagonal constrained scenario we require in addition consistency of $\V(x_i,x_j)$ for all coupled pairs of variables. 

The quantities made consistent are written more concisely as
\begin{eqnarray}
  \V_p(x_{i},x_{j}) &\approx& C_{i,j} \defeq \V_{q_{(ij)}}(x_i,x_j) \\
  \V_p(x_{i},x_{j}) &\approx& \chi_{i,j} \defeq \sum_{x_i,x_j} q^*_{(ij),i}(x_i,x_j) x_j
\end{eqnarray}
When evaluated at minima of the free energy, both the approximation by marginalization
$C^*$, and approximation by linear response $\chi$, are symmetric.

\section{Minimizing with respect to the constraints}
\label{sec:freeenergyminimization}

We choose a Lagrangian formulation for minimizing the constrained free energy, introducing
a Lagrange multiplier for each constraint connecting a linear response approximation to a
marginalization approximation. Each statistic pair constraint will be associated with some
unique outer region $\alpha$, as this allows for a cavity heuristic that we later introduce. This
association is not unique and may affect (to a limited extent) the convergence of the
algorithms we will develop, but not the fixed points that might be achieved. This is discussed
further in Appendix \ref{app:constraintAssignment}. The set of constraints associated to region $\alpha$ are denoted $\boldsymbol{\omega}_\alpha=\{[(i_1,y_1),(i_2,y_2)]\}$ with an associated set of Lagrange multipliers $\boldsymbol{\lambda}_\alpha = \{\lambda_{(i_1,y_1),(i_2,y_2)} \}$. The constrained minimization is then achieved by minimizing the Lagrangian
\begin{equation}
  F_\nu(q,\lambda,\chi) = F_\nu(q) + \sum_\alpha \left[\sum_{[(i_1,y_1),(i_2,y_2)] \in \boldsymbol{\omega}_\alpha} \lambda_{(i_1,y_1),(i_2,y_2)} \Delta_{\alpha,(i_1,y_1),(i_2,y_2)}(q_\alpha,q^*_{\alpha,(i_1,y_1)}) \right] \label{Lagrangian}
\end{equation}
with Lagrange multipliers set to meet the constraints.
The global minima $q^*$ for fixed $\lambda$ can often be found; in cases where local minima can be avoided this is done either heuristically following a constrained loopy belief propagation (CLBP) approach or in a more robust manner using a provably convergent method, as discussed in Appendix \ref{app:Algorithms}. CLBP has
the same (asymptotic) computational complexity as belief propagation, and is procedurally
similar.
\begin{figure}
\centering
  \includegraphics[width=\linewidth]{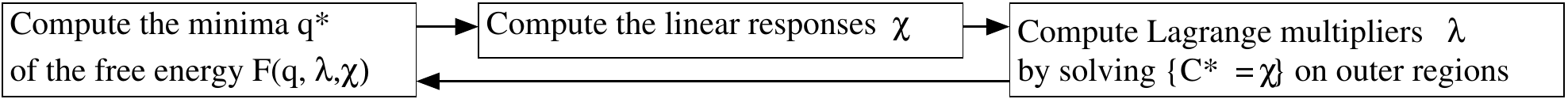}
  \caption{\label{fig:3cycle} The basic scheme for minimizing our constrained variational free energy. The first stage is solved by a procedure closely related to belief propagation (that minimizes a convex-concave function), the second by a procedure related to susceptibility propagation (that solves a system of linear equations), and the final stage by a cavity approximation. In our experiments, we take $\lambda=0$ as the initial condition. In the experiments presented we gradually increase or decrease $T$ using the solution at $T \pm \delta T$ as an initial condition for the next experiment, this enables a solution to evolve continuously from a well-understood limit, but is not necessary for convergence in general.}
\end{figure}

It is shown in Appendix \ref{app:linearresponse} that given such a fixed point, whether a local or global minimum, it is subsequently relatively easy to calculate the linear response. The method we propose is procedurally similar to of susceptibility propagation, originally introduced in ~[\cite{Mezard:CSPNN}], which
is a computational procedure that minimizes the variational parameters; if the number of
constraints is linear in the size of the system the computational complexity is quadratic in
system size; we call the linear response scheme constrained loopy susceptibility propagation
(CLSP). Yasuda et al. proposed a closely related approach, specific to the case of diagonal constraints in the Bethe approximation~[\cite{Yasuda:SPDC}].

Suppose we also have a method for iteratively determining $\lambda$, then we can approach the problem of finding a constrained local minima by a 3-stage iterative procedure (the same as proposed in~[\cite{Yasuda:SPDC}]), and shown schematically in Figure \ref{fig:3cycle}.  We also experimented with other minimization schemes, but found this to be algorithmically the most stable, and also pleasing in that we move all uncertainty in convergence of the method onto the two questions: (1) does there exist a fixed point at all and (2) does the iterative scheme for $\lambda$ converge. 
These two questions are unfortunately very difficult to answer in general. 
To determine $\lambda$ a heuristic {\em cavity} method is proposed in Section \ref{cavityScheme}.

We found that in weakly correlated regimes the CLBP and CLSP, and iterative updating of $\lambda$ quickly converges. 
However, in some other regimes where the variational approximation is less accurate, we found that the effect of non-zero $\lambda$ could be either a help or a hindrance to the convergence of the CLBP. We also found that $\lambda$ would sometimes not converge, even with strong damping (implementation of damping is discussed in Appendix \ref{app:solvingforlambda}). 
Associated with regimes of non-convergence, we normally find a divergence of some $\lambda$ values as discussed in the experimental section. 
This indicates that failure of the method is most likely related to non-existence (or criticality) of constrained solutions, the issue of solution existence is discussed in Appendix \ref{app:uniqueness}.

\subsubsection{Bethe approximation to the Boltzmann machine}
For the on-and-off diagonal constrained case, the Lagrangian for the Boltzmann machine can be written in the form\footnote{To maintain consistency with published research on Ising spin models a factor $1/2$ precedes the diagonal constraint term.}. 
\begin{multline}
  F_\nu(\boldsymbol{q},\boldsymbol{\lambda},\chi) = \sum_{(i,j)\in E}\sum_{x_i,x_j} q_{ij}(x_i,x_j) \log q_{ij}(x_i,x_j) + \sum_i (1-k_i) \sum_{x_i} q_i(x_i) \log q_i(x_i) 
  \\ + \frac{1}{2} \sum_{i} \lambda_{i,i} \left[ (1 - M_i^2) - \chi_{i,i} \right]
   + \sum_{(i,j) \in E} \lambda_{(i,j)} \left[ \sum_{x_i,x_j} q_{ij}(x_i,x_j) x_i x_j - M_i M_j - \chi_{i,j} \right] \label{eq:Lagrangian2}\;,
\end{multline}
introducing abbreviations for single variable magnetization $M_{i} \defeq \sum_{x_i} q_{i}(x_i) x_{i}$.
We recover the diagonal constraint regime when $\lambda_{i,j}=0\qquad \forall i,j$, and the unconstrained regime when all multipliers are zero.
Constraints introduce quadratic functions of $q$, but terms are neither convex nor concave.

For given $\lambda$, a set of message passing equations can be written as a generalization of loopy belief propagation, a simplification of the general case in Appendix \ref{app:pseudocode} is presented here. At time $t$ the edge-belief is approximated as the solution to the equation 
\begin{multline}
  q_{ij}^t(x_i,x_j) \propto  \mu^t_{i \rightarrow (i,j)}(x_i)\mu^t_{j \rightarrow (i,j)}(x_j) \\ \exp\left[(J_{ij}-\lambda_{ij}) x_i x_j + (h_i + \lambda_{ij} M^t_{ij,j} + \lambda_{i} M^t_{ij,i}) x_i + (h_j + \lambda_{ij} M^t_{ij,i} + \lambda_{j} M^t_{ij,j} )x_j\right] \label{eq:qt}
\end{multline}
where we introduce two auxiliary magnetization parameters per edge\footnote{Note that at intermediate stages of the message passing, magnetizations for variable $i$ on different beliefs (say $M_{ij,i}$, $M_{ik,i}$) may not agree, but will agree after convergence.}
\begin{equation}
  M_{ij,i}^t = \sum_{x_i x_j} x_i q_{ij}^t(x_i,x_j) \label{eq:Mt}\;.
\end{equation}
 Then we can define messages\footnote{These differ by a simple transformation from the
   messages of the Appendix \ref{app:pseudocode}, so that the limiting case ($\lambda=0$) agrees with standard presentations for Ising models in the literature.} $\mu$ which are determined iteratively as
\begin{eqnarray}
  \mu_{(i,j) \rightarrow i}^t(x_i) &\propto& \sum_{x_j} \mu^t_{j \rightarrow (i,j)}(x_j) \nonumber \\
&&\qquad \exp\left[(J_{ij}-\lambda_{ij}) x_i x_j + (h_j + \lambda_j M_{ij,j}^t + \lambda_{ij} M_{ij,i}^t) x_j +  \lambda_{ij} M_{ij,j}^t x_i \right]  \nonumber\\
  \mu^{t+1}_{i \rightarrow (i,j)}(x_i) &=& \prod_{k \in \partial_i \setminus j} \mu_{(i,k) \rightarrow i}^t(x_i) \label{eq:mu}
\end{eqnarray}
where $\partial_i$ are the variables interacting with $i$. 
Following message updates $q^t$ and $M^t$ must be made consistent, in the examples of this paper (and in general for small $\lambda$) this can be achieved simply by iterating equations (\ref{eq:Mt}) and (\ref{eq:mu}). 
For the experiments messages and beliefs are initialized as constants.

There are several methods by which linear response can be established. One standard approach is susceptibility propagation, which simply involves linearizing the above equations accounting for a small perturbation in some component of $\nu$, this is the approach taken for the examples of this paper; for the general case, expressions are provided in Appendix \ref{app:linearresponse}. 

\subsection{Determination of $\lambda$}
\label{cavityScheme}
To determine $\lambda$ we propose the following scheme. We wish to solve at each $\alpha$ a set of non-linear equations $\{\boldsymbol{\Delta}_{\alpha}(q_\alpha,q^*_{\alpha,(\cdot)})=0\}$, where $q_\alpha$ and $q^*_{\alpha,(\cdot)}$ are functions of all $\boldsymbol{\lambda}$ (through the system of message passing equations). One possibility is to linearize these equations about the current estimate (i.e. apply Newton's method), but this leads to an impractical $O(N^3)$ procedure, dominating other algorithmic time-scales for moderately sized systems. 
%Some related procedures like rank-one update methods also prove to be expensive.

Instead, we resort to a locally consistent and parallelizable approximation: 
For some $\boldsymbol{\lambda}$ we find a minimum defined by messages
$\boldsymbol{\mu}$ (Appendix \ref{app:pseudocode}), and the linear response for these messages (Appendix \ref{app:linearresponse}). 
If $\boldsymbol{\lambda}$ is approximately correct
then the messages passing into
the region $\alpha$ should be weakly dependent on any changes to $\boldsymbol{\lambda}_\alpha$ in that region. 
Since we can define $\boldsymbol{\Delta}_\alpha$ in terms of the local parameters
$\boldsymbol{\lambda}_\alpha$, and the incoming messages (which we argue are unchanged by an update of $\boldsymbol{\lambda}_\alpha$)
the problem for determining $\boldsymbol{\lambda}$ is reduced to solving for $\boldsymbol{\lambda}_\alpha$ independently on every region.

\begin{figure}
\centering
  \includegraphics[width=0.5\linewidth]{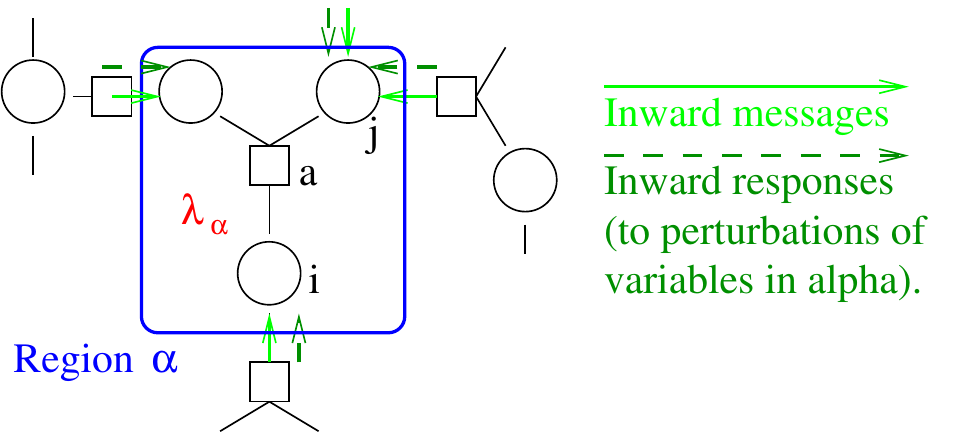}
  \caption{\label{fig:cavityargument}  Given $C^*$ and $\chi$, determination of $\lambda$ breaks into a set of independent problems on outer regions $\alpha$. 
    This can be understood as a cavity approximation.
    Assuming the incoming messages, and responses,
    to be fixed and approximately independent of the Lagrange multipliers to be set
    ($\boldsymbol{\lambda}_\alpha$), we can independently, on each region, solve the system of equations
    $\boldsymbol{\Delta}_\alpha=0$ for $\boldsymbol{\lambda}_\alpha$. A similar assumption often justifies message passing and mean-field iterative heuristics. }
\end{figure}

In this way, we arrive at a cavity-approximation style argument common
in the motivation of message passing algorithms, see Figure
\ref{fig:cavityargument}. However, it is noteworthy that unlike
LBP, incoming messages and responses depend on $\boldsymbol{\lambda}_\alpha$
locally. There is a direct feedback that exists even in the absence of loops. 

Unfortunately, $\boldsymbol{\Delta}_\alpha=\mathbf{0}$ remains a (small) system of non-linear equations in
$\boldsymbol{\lambda}_\alpha$, that does not allow a closed form solution in general. 
One exception is when only diagonal constraints are applied~[\cite{Yasuda:GISP,Yasuda:SPDC}], and the cavity argument is applied to single-variable regions. More generally we
use Newton's method to solve these equations
\begin{equation}
  \left.  {\left[\frac{\partial \boldsymbol{\Delta}_{\alpha}(q^*,q^*_{(\cdot)})}{\partial \boldsymbol{\lambda}_\alpha} \right]^{-1} \boldsymbol{\Delta}_{\alpha}} \right|_{\lambda=\lambda^t} \delta \boldsymbol{\lambda}_\alpha^{t+1} = - \left. \boldsymbol{\Delta}_{\alpha}(q^*,q^*_{(\cdot)}) \right|_{\lambda=\lambda^t} \label{eq:iteration}\;.
\end{equation} 
where the dependence of $q^*$ and $q^*_{(\cdot)}$ on $\lambda$ follows from (\ref{eq7}) and (\ref{eq:dqstar}) with messages fixed.

An alternative expression based on a closed form for the covariance matrix approximation $\chi$ is discussed in  Appendix \ref{app:solvingforlambda}, alongside some other technical details on the determination of $\lambda$.

\section{Results}
\label{sec:results}
In this section, we study the performance of our method on
well-understood toy model frameworks. The scale and/or symmetries of these models mean they are exactly solvable, allowing precise statistical estimates to evaluate the method quality.

In Section \ref{sec:FC} we study a fully connected ferromagnetic model (i.e.\ a model with a positive coupling between any pair of variables) with symmetry broken (that is with a non-zero mean value for each variable).
This is a simple model for which we can present analytic results and understanding. There is
either no mode (for weak coupling), or one dominating mode (for strong couplings). We expect
the method to be weakest for intermediate coupling strength since the Bethe approximation
becomes exact (with corrections $O(1/N)$) in the limit of strong (a single mode) or weak (no modes) coupling. In Section
\ref{sec:WJ} we study a model with frustration (that is couplings have different signs and it is not possible to find a configuration satisfying them all at the same time) and a random distribution of optima and sub-optima.
The problem is multi-modal in the limit of strong
couplings and the Bethe approximation breaks down. In Section \ref{sec:Potts} we consider a simple
model with an expanded discrete alphabet, where randomness is introduced through a random
graphical structure. Like the ferromagnetic example, a single mode dominates for strong couplings. We anticipate the Bethe approximation to become exact in the
limit of large problems, but for smaller problems the presence of short loops leads
to inaccuracy. In the final example of Section \ref{sec:alarmNet} we consider a well-studied toy model
involving both multi-variable interactions and multi-states, both the
Bethe approximation and linear response perform poorly on this model. These examples cover
a range of scenarios in which our method might be applied. Special cases of the variational
approximation we present have previously been applied to lattice models commonly studied in statistical physics,
and sparse prior models in Bayesian image modeling~[\cite{Raymond:MFM,Raymond:Correcting,Yasuda:SPDC,Yasuda:GISP}].

We present behaviour of the Lagrange multipliers $\lambda$,  the self-consistency error (\ref{eq:Delta}), errors on
the pair statistics when using either marginals (\ref{eq:belieferror}) or linear
responses (\ref{eq:linearresponseerror}), and errors on the marginal estimates
\begin{equation}
  \Delta^{(0)}_{(i,y)}(q^*) = q^*(x_i=y) - p(x_i=y) \label{eq:Marginalerror}
\end{equation}
The maximum absolute deviation (MAD) on the marginals is defined as
the largest error over all variables, or pairs of variables, depending on error type.

It is interesting to understand how the quality of approximation
changes as a function of the goodness of the approximation, to do this
we introduce a temperature parameter $T$ in each model, that can sharpen or flatten the distribution.
\begin{equation}
  p_T(x) \propto p(x)^{1/T}
\end{equation}
We can minimize with relative ease both the constrained and unconstrained free energies in the large $T$ regime, and expect approximations to be correct at leading order in $1/T$~[\cite{Raymond:Correcting}]. In some cases of small $T$, in which the probability is well described by a single mode, concentrated about some unique value $x_{GS}=\mathrm{argmax}\;p_T(x)$, the approximations we present are also exact up to $O(T)$ corrections.

We find that in many of the models, minimization of the
constrained free energy is slow or impossible for
$T$ over some intermediate range, or below some threshold. To
extend the range of $T$ for which solutions could be found, an annealing procedure was employed: beginning at
large (or small) $T$ and proceeding through a sequence of models
slowly changing $T$ - and using the solution to the previous model as the initial condition for the subsequent minimization.
Under this procedure, we find that $\lambda(T)$ evolves smoothly, but
that there appears still to be a limit in the accessible temperature range. In the applications presented we did not find fixed points that appeared discontinuously - it seems that either a solution can be reached by annealing from low temperature, from high temperature, or is absent.
 
Our motivation for introducing $T$ is threefold: to study the breakdown of the approximation (absence of solutions), to mitigate for non-convergence, and to increase the speed of convergence. The annealing procedure introduces additional computational costs - we have not made timing comparisons against loopy belief propagation or other competitors, for either the simple, or annealed, procedure. Instead, we have prioritized an exploration
of the nature of solutions that can be discovered, and we have sought very accurate estimates to the parameters describing those solutions. Compromises in the accuracy of constraint
satisfaction, annealing rate (or absence of annealing), and damping can all have a significant
impact on the speed of the method.

\subsection{The fully connected ferromagnet}
\label{sec:FC}
\begin{figure}
  \includegraphics[width=0.49\linewidth]{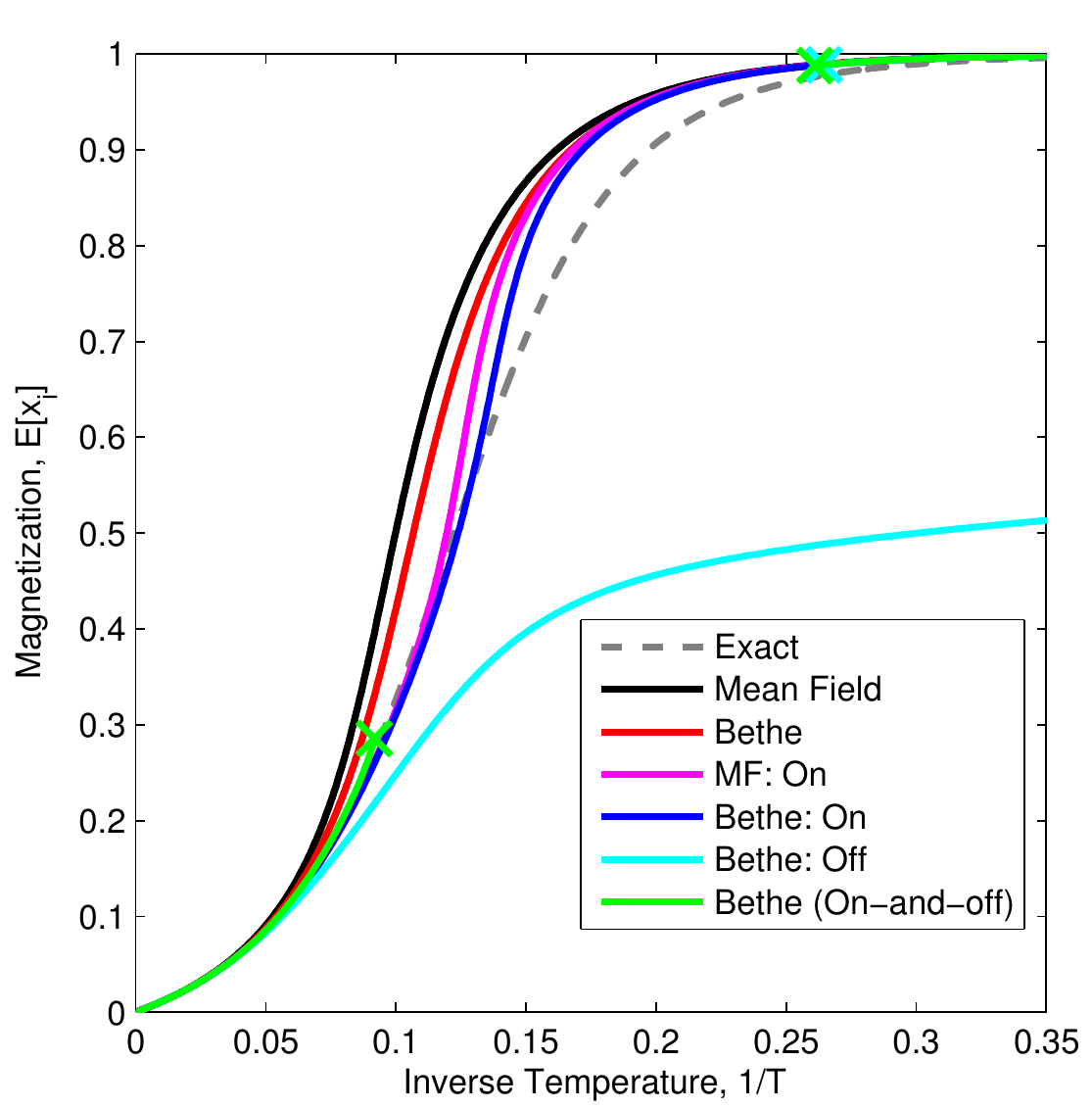}
  \includegraphics[width=0.49\linewidth]{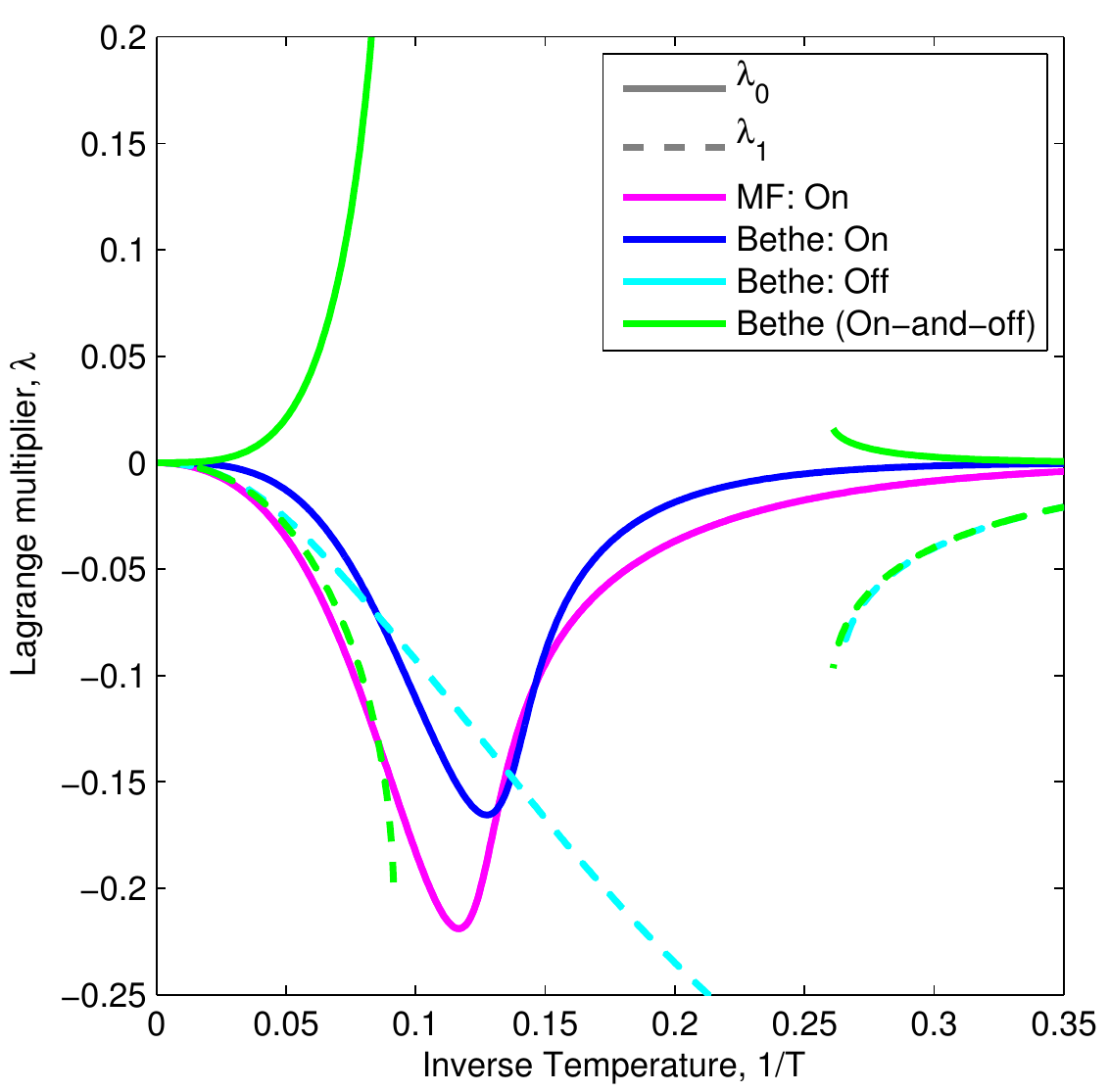}
\caption{\label{fig:N10} Results for the fully connected
  model (\ref{eq:fullConn}) with $N=10$ and $h=1$. Results are
  presented for the mean-field and Bethe approximation schemes,
  with and without constraints. (left)
  The magnetization is overestimated in the Bethe and Mean-field approximations, the effect of the constraints is to suppress the magnetization . Certain solutions exist only in the high-temperature or low-temperature regimes, those solutions are demarcated by 'x'. 
  One solution to the Bethe approximation with off-diagonal constraints only
  suppresses the magnetization too strongly, there is another solution that is present only at low temperature, where the estimate is more reasonable. 
  Where it exists, the Bethe approximation with on and off-diagonal constraints is the most accurate.
  (right) There are at most two independent values for
  $\lambda$ in this model: diagonal $\lambda_0$ and off-diagonal $\lambda_1$. The Lagrange multipliers deviate most from zero
  at $T\sim 1/N$ (this is related to a ferromagnetic phase transition). $\lambda$ diverge rapidly in some solutions demarcating regions where no solutions can be found. }
\end{figure}
We begin with a simple but informative case, that of a fully connected
ferromagnetic Ising model in an external field. 
\begin{equation}
  p_T(x) = \frac{1}{Z} \exp\left[\frac{1}{2 T}\left(h + \sum_{i=1}^N x_i
    \right)^2\right] \label{eq:fullConn}
\end{equation}
Variables are Ising spins $x_i = \pm 1$.
The marginals for this model can be solved
up to $O(1/N)$ by the mean-field approximation, and up to $O(1/N^2)$ by the Bethe
approximation. Accuracy is a function of temperature when this
is large or small compared to $1/N$ the accuracy is correspondingly
high. Under our method the equations to be solved, and
associated errors, can be expressed concisely; this is done in
Appendix \ref{app:FCIsing}.

Three kinds of constraint are considered: 
diagonal $\{\Delta_{ii}=0,\; \forall \; i\}$;
off-diagonal $\{\Delta_{ij}=0,\; \forall i\neq j\}$; and on-and-off diagonal applying both sets. Results for both the mean field and Bethe approximations are presented. Only diagonal
constraints can be applied in the mean field case since the variational parameters are consistent only with zero off-diagonal covariances.

The left panel in Figure \ref{fig:N10} shows the behavior of the magnetizations, $\E(x_i)$ 
for the case $N=10$ and $h=1$. The most accurate estimates
are obtained with on and off-diagonal constraints applied at the Bethe approximation at high and low
temperature. The unconstrained approximations overestimate the magnetization (bias in variables). The addition of 
constraints corrects this bias: in one constrained regime (off-diagonal constraints only) the suppression
is clearly too strong at intermediate $T$, while in all other cases effective. 

In some constrained approximations there is a smooth evolution of the optimal beliefs with $T$, which
reflects the behavior of the exact marginals, in others, the optimal beliefs for different $T$ are not continuously related.

When off-diagonal constraints are applied in the Bethe approximation, either alongside
or without diagonal constraints, we see a discontinuous emergence of the strongly magnetized solution. In the case of only off-diagonal constraints, there is a coexistence of two fixed
points for small $T$. This means that as we vary the parameter $T$ the $q^*$ moves discontinuously from a relatively smooth approximation to
one characterized by a single mode. As $N$ increases the range over which coexistence exists
shrinks, and all approximation approach the correct result for large $N$.

With on and off-diagonal constraints we find a range of $T$ for which no solutions can be
found by a continuous evolution of the low or high-temperature solutions. It seems highly
likely that no solution exists, and empirically we were not able to find fixed points (for any
model) that were not continuously related to either a high or low-temperature solution.

The behaviour of $\lambda$ (see right panel in Figure \ref{fig:N10})
and indeed a careful examination of the
fixed points indicate why solutions disappear in this simple
case. The values of the Lagrange multipliers diverge, and this is
related to marginals approaching their boundary values - where
variational inference breaks down due to the inflexibility of the
parameters.
Decreasing $h$ or $N$ we increase the inaccuracy of the Bethe
approximation at intermediate $T$, this can lead to discontinuity also
for the diagonal constrained solution, and reduces the range of
temperatures over which the low-temperature solution (the one with a large magnetization)
can be found for constrained problems.

%The case of lattice models with uniform interactions was studied in
%\cite{Raymond:MFM,Raymond:Correcting}: owing to symmetries we were
%able to make simplifications, and observed a similar variety of behaviors to th%ose presented here.

\subsection{The Wainwright-Jordan set-up}
\label{sec:WJ}
A common toy model on spin variables $x_i=\pm 1$ is the Wainwright-Jordan set-up where $N = L \times L$ Ising spins are arranged on a square grid~[\cite{Opper:IEP}].
\begin{equation}
  p_T(x) = \prod_{(i,j) \in E} \exp(J_{ij} x_i x_j/T) \prod_i \exp(h_i x_i/T) \;.
\end{equation}
Fields $h_i$ are independent and identically distributed (iid) samples from $[-0.25, 0.25]$ and couplings $J_{ij}$ are sampled i.i.d on $[-1, 1]$. The Bethe
approximation fails on these models for smaller T owing to multi-modality of the distribution, but for
larger T (where correlations are weaker) the approximation is a significant improvement
upon the mean-field approximation.
 
Diagonal $\{\Delta_{i,i}=0, \;\forall \; i\}$, or diagonal and 
off-diagonal $\{\Delta_{i,j}=0,\forall (i,j) 
\in E\}$ constraint regimes are studied for the Bethe approximation.

\begin{figure}
\includegraphics[width=0.49\linewidth]{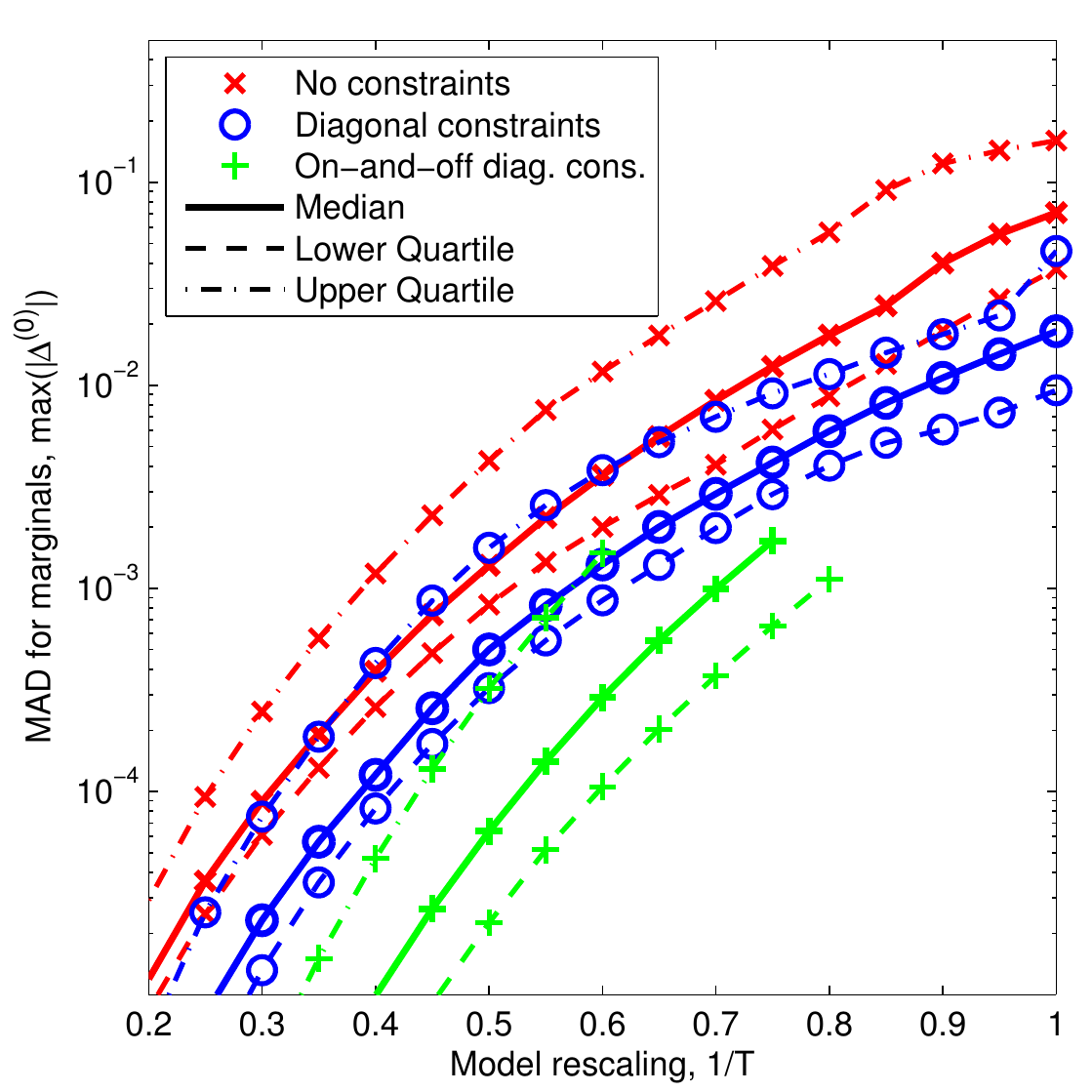}
\includegraphics[width=0.49\linewidth]{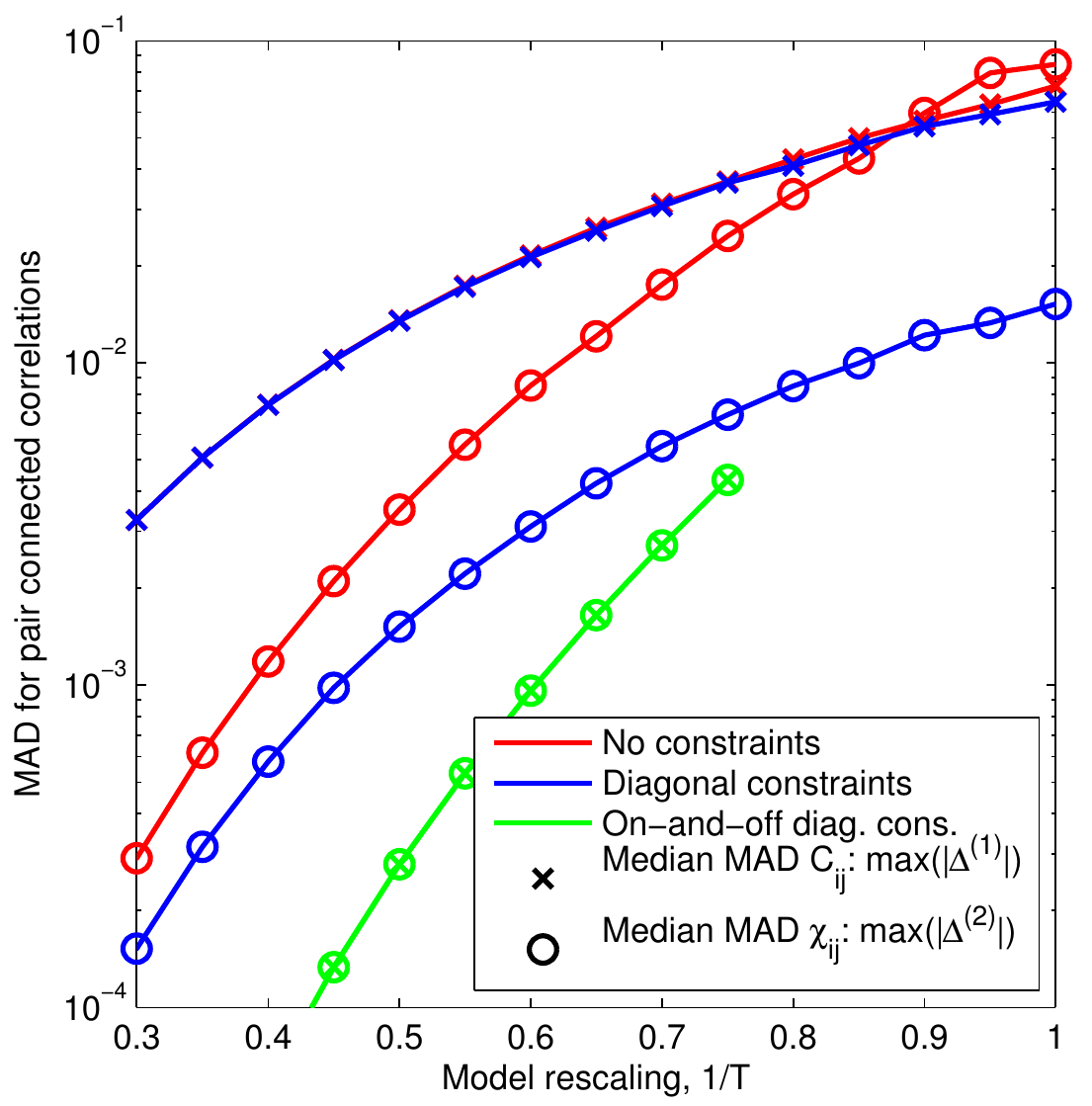}
\caption{\label{fig:JordanWainwright4} $L=4$ Wainwright-Jordan set-up:
  The error on the marginals, and connected correlations (which
  together provide a sufficient description of pair probabilities) are
  improved everywhere by adding constraints, as long as the method
  converges. As discussed, the MAD for $C_{ij}$ is worse than for
  $\chi_{ij}$, although they are becoming comparable approaching $T=1$. On the right the median over 20 models is shown, on the left the median and quartiles. The advantage is consistent across all models.}
\end{figure}

\begin{figure}
  \includegraphics[width=0.49\linewidth]{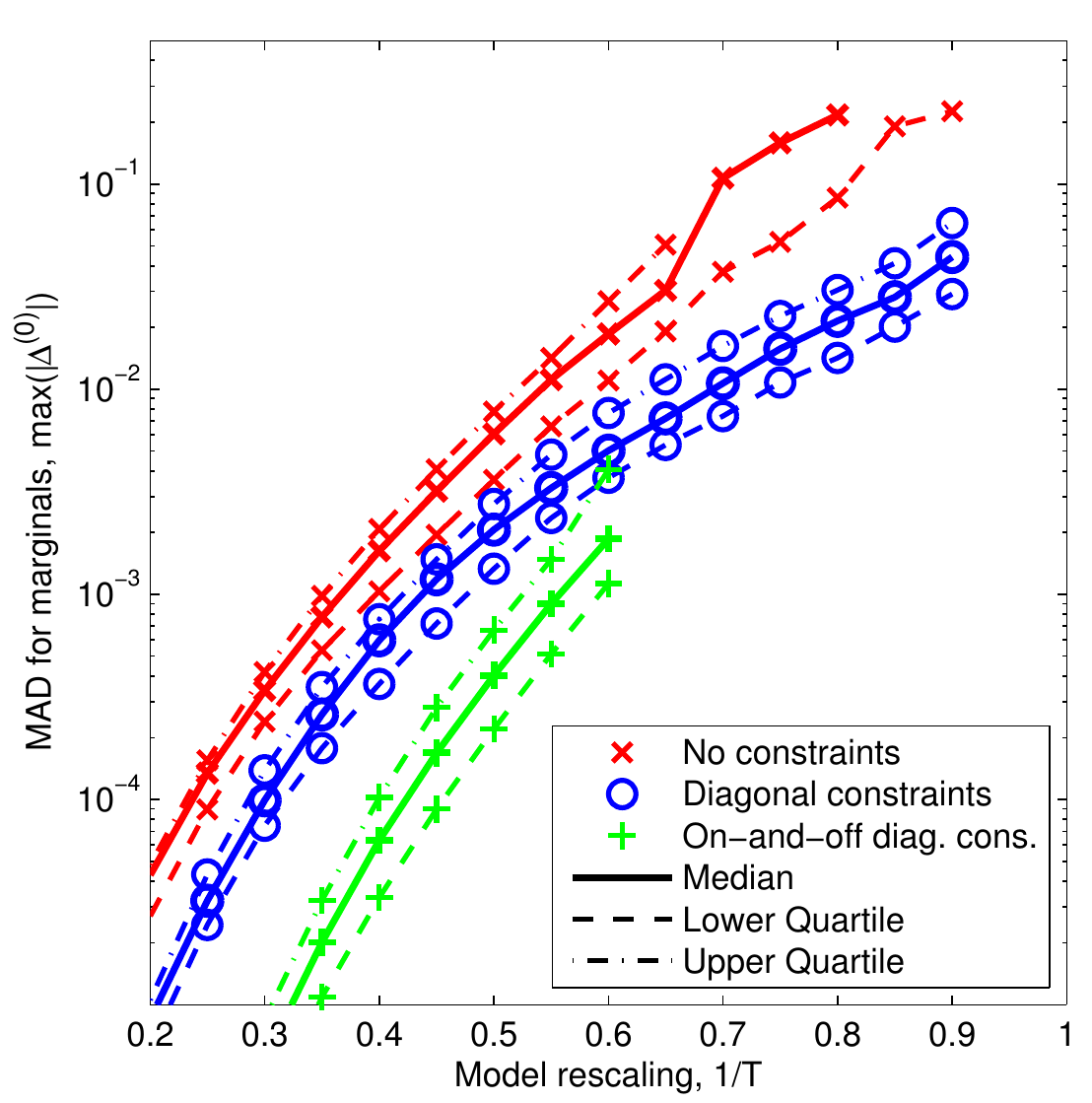}
  \includegraphics[width=0.49\linewidth]{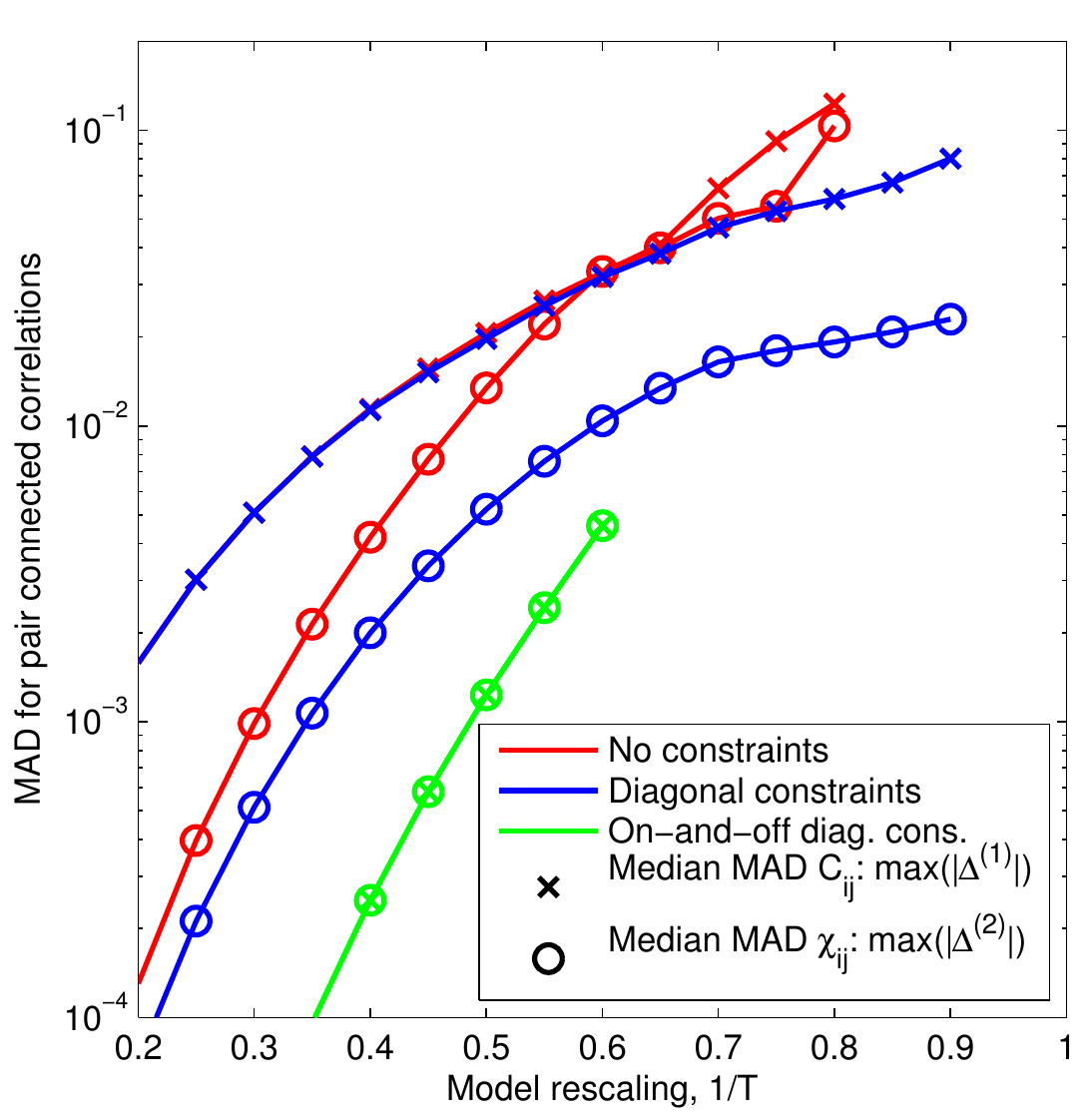}
\caption{\label{fig:JordanWainwright7} $L=7$ Wainwright-Jordan set up: Trends are comparable to the smaller system in Figure \ref{fig:JordanWainwright4}, where solutions exist significant gains are made in all models with the addition of constraints. However, all approximations are now failing to reach full scale. The method which is stable to lowest temperature is the model with diagonal constraints only, while the one with on-and-off constraints is the most accurate.}
\end{figure}

We considered 20 instances for $L=4$ and $L=7$, and plot the MAD
results in Figures \ref{fig:JordanWainwright4} and
\ref{fig:JordanWainwright7}. Shown are the quartiles, 
for cases where methods did not converge we
assigned value +Inf to the MAD so that the quartile values are
truncated in cases where the fraction of non-convergent cases exceeded
the quartile. 
For the smaller system ($L=4$) the Bethe approximation succeeds for typical
cases to full scale ($T=1$) either without
constraints or with only diagonal constraints; the on-and-off
diagonal scheme fails in a larger fraction of cases. 
For the larger $L=7$ system LBP and CLBP fail at full scale: the model 
which prevails to lowest temperature is CLBP with diagonal
constraints; the model failing soonest is CLBP with on and off-diagonal constraints.
The exact range of $T$ for which methods were convergent was sensitive to
the annealing procedure, the amount of damping used, and the convergence criteria; strong
damping and slow annealing broaden the range for all
methods.

Where the algorithm is convergent the performance pattern is very
clear - there is significant improvement adding diagonal constraints,
and more so with on and off-diagonal constraints.
The quartiles demonstrate that the MAD advantage is maintained across a broad range of models. In examining the detailed distribution of improvements, we find it is not only the maximum deviation that is significantly improved, but the mean, and almost every individual statistic of the model. For $N=16$, we can compare the median behavior against that reported in ~[\cite{Opper:IEP}]; where the strongest method (tree-EP) also improves the MAD result for marginals by approximately one order of magnitude. Expansion methods offer some further (but modest gains)~[\cite{Paquet:PC,Opper:PC}].
  
On-diagonal constraints allow an increased range (in $T$) for convergence. Qualitatively,
we offer the following explanation. In the constrained regimes (in the high-temperature regime, the only one for which we demonstrate solutions) the majority of Lagrange multipliers are negative. The effect is to suppress biases and decrease susceptibility (the sensitivity of biases to small changes in the parameters). We assume that reduced susceptibility also correlates with reduced sensitivity to small fluctuations in the messages, which should aid algorithmic stability.
With on-and-off diagonal constraints the diagonal and off-diagonal Lagrange multipliers are strongly dependent where they constrain the same variable. The off-diagonal
multipliers are typically negative, and effectively suppress biases, by contrast, the diagonal
multipliers are positive and reinforce biases. The typical net effect is to reduce biases and susceptibility, as with the on-diagonal case. However, the strongly correlated nature of the parameters may be the source of convergence problems.

\begin{figure}
  \includegraphics[width=0.49\linewidth]{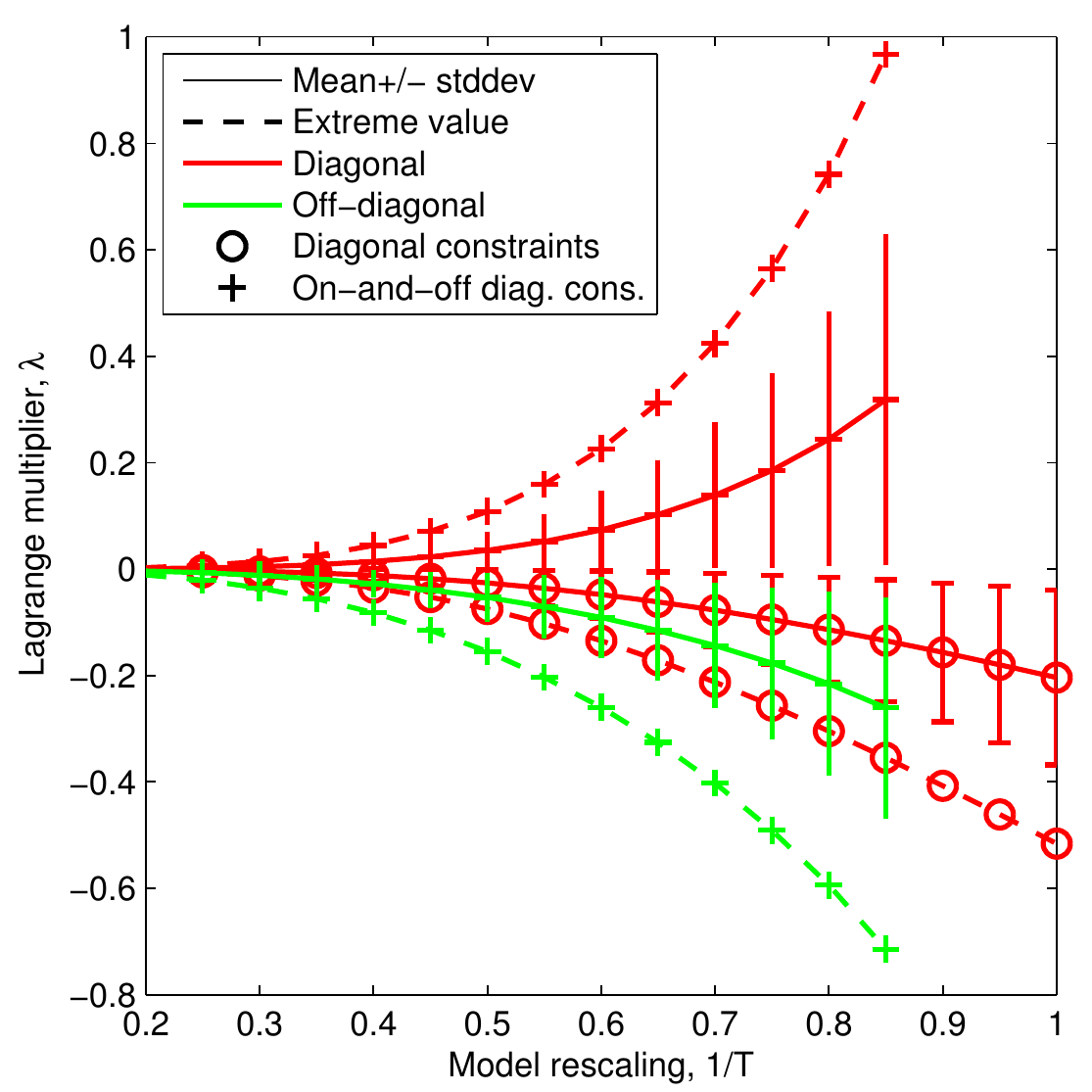}
  \includegraphics[width=0.49\linewidth]{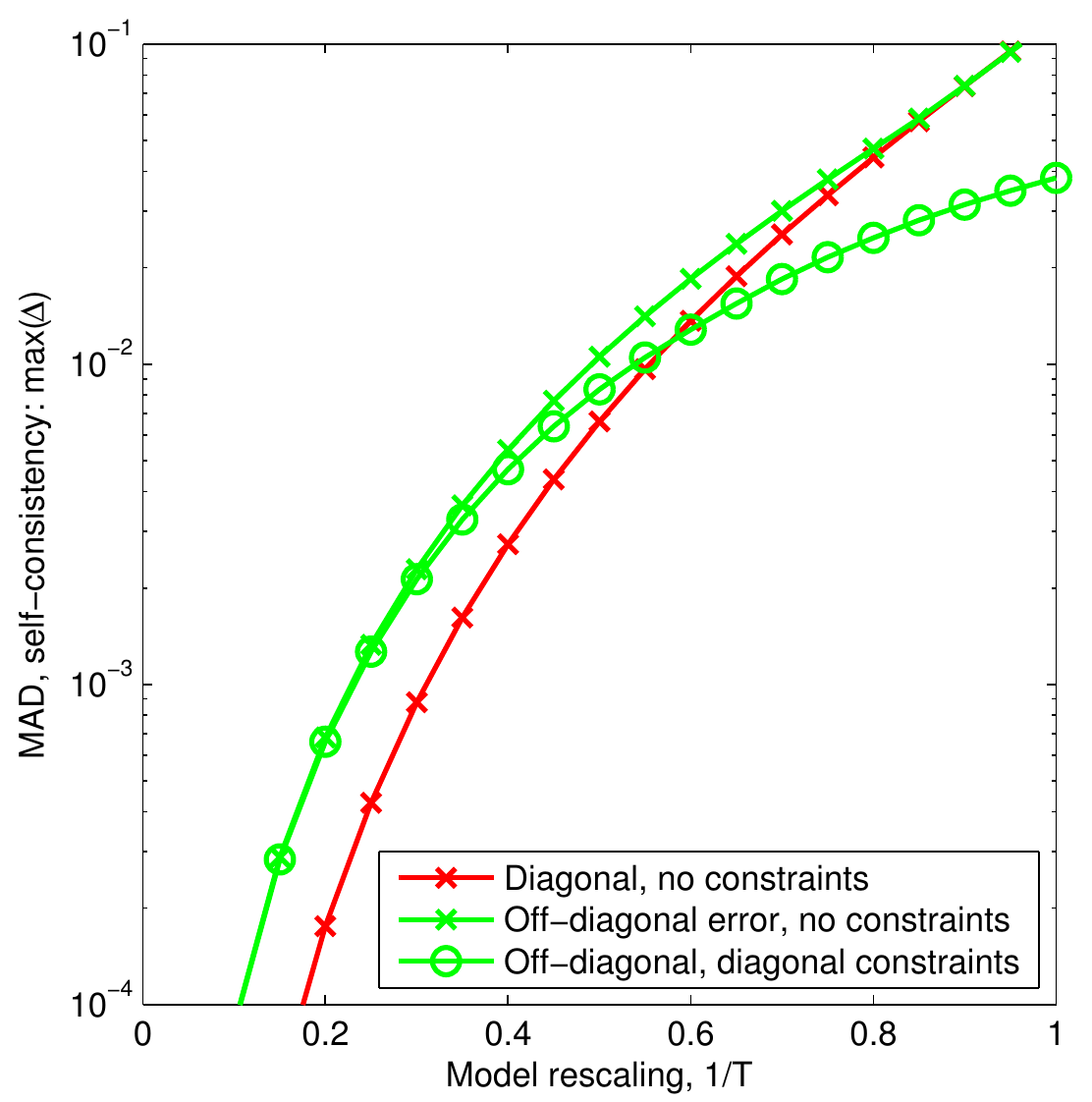}
\caption{\label{fig:JordanWainwrightLambda} A typical problem in the
  Wainwright Jordan set up for $N=16$. (left) Negative $\lambda$ has the effect of effectively reducing the coupling strength; as with the fully connected ferromagnetic model, a mode of failure for the approximation is to over-estimate the bias in variables, and constraints appear to work by mitigating this effect. Bars indicate the 3 quartiles of the distribution, and the extremal values are also plotted. In the Bethe
  approximation with only diagonal constraints, most $\lambda$
  including the extremal value are negative.  With on and off-diagonal constraints $\lambda$ with a net effect of reducing interaction strength, values diverge approaching the point
  of algorithmic failure. (right) Adding diagonal constraints not only removes the diagonal self-consistency
  error, but also reduces the error in the unconstrained (off-diagonal)
  statistic.
   Adding both on and off diagonal constraints removes inconsistency for both types. The self consistency errors (\ref{eq:Delta}) grow with decreasing temperature.  
}
\end{figure}

The failure of CLBP is most often due to non-convergence of $\lambda$, rather than
a failure in the CLBP or CLSP iterative algorithms (at fixed $\lambda$). Figure \ref{fig:JordanWainwrightLambda} indicates why
the iterative update is failing: some Lagrange multipliers are diverging in a strongly correlated manner and it seems likely there
is a critical value of $T$ close to the failure point beyond which no solutions exist, as found in Section \ref{sec:FC}.

The LBP implementation follows the same procedure of Appendix \ref{app:Algorithms},
as the constrained cases, with the difference that the innermost do--while loop is always
convergent in one iteration, and $\boldsymbol{\lambda}=0$. Using a double loop procedure we might force
convergence to a minimum of the Bethe approximation, but the convergence properties of LBP are still an interesting point of comparison.

We might seek to extend the range of convergence for CLBP by clever modifications.
However, it seems that breakdown of convergence is closely related to breakdown of the
underlying (Bethe) approximation. Algorithmic innovations would not extend significantly
the range of problems for which the constrained approximation is useful in practice; just as
the availability of double loop methods has not revolutionized the use of the Bethe approximation: where LBP fails the Bethe approximation is almost always a poor
approximation.

For this type of problem a significant improvement to the Bethe approximation is made in moving  to a plaquette-based Kikuchi approximations for the case of regular lattices~[\cite{dominguez2011characterizing,lage2013replica}]. Limited experiments on constrained Kikuchi approximations also indicate a modest decrease in error on marginals with the application of constraints.


\subsection{Potts model in an external field}
\label{sec:Potts}

 \begin{figure}
\includegraphics[width=0.49\linewidth]{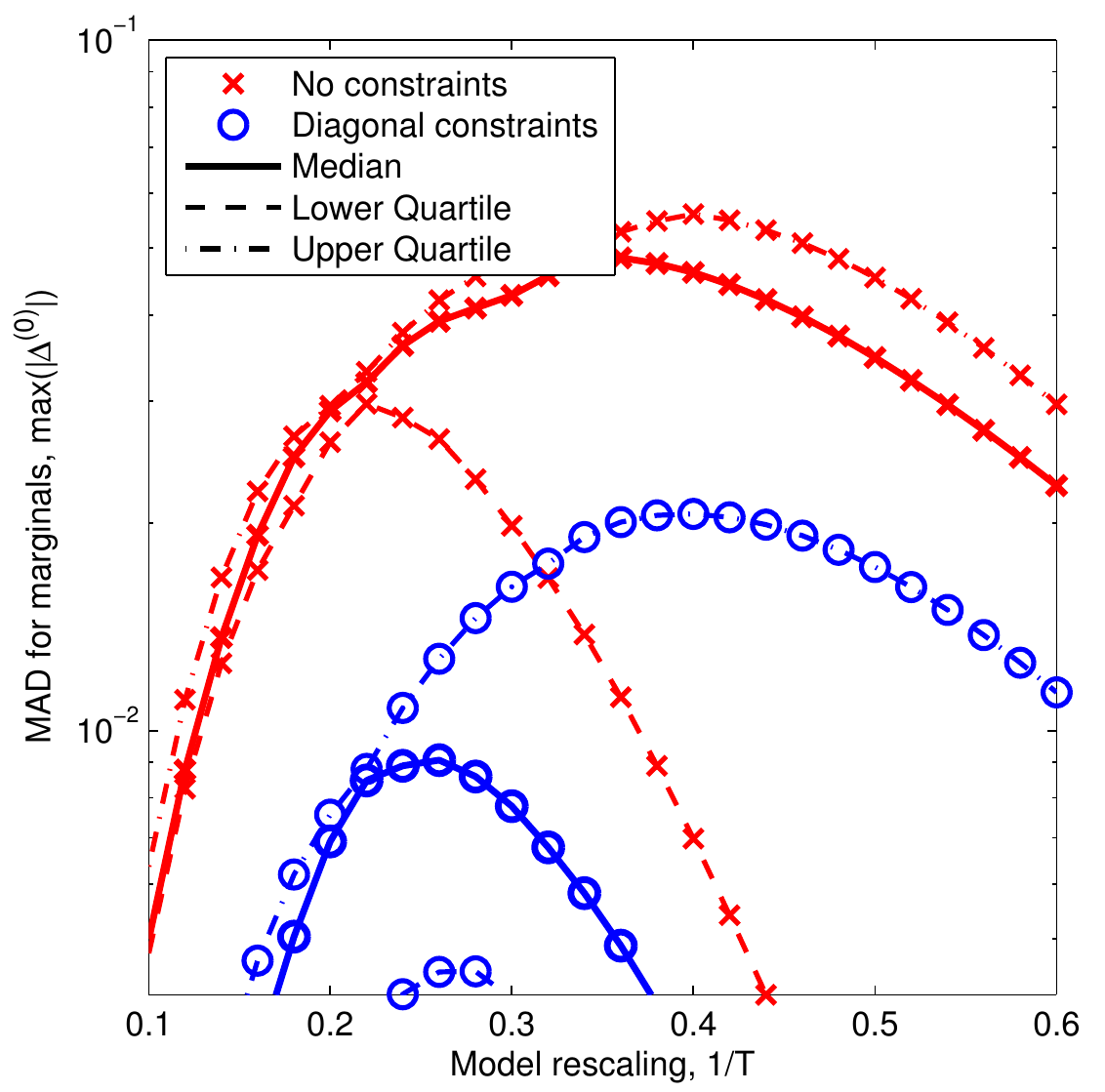}
\includegraphics[width=0.49\linewidth]{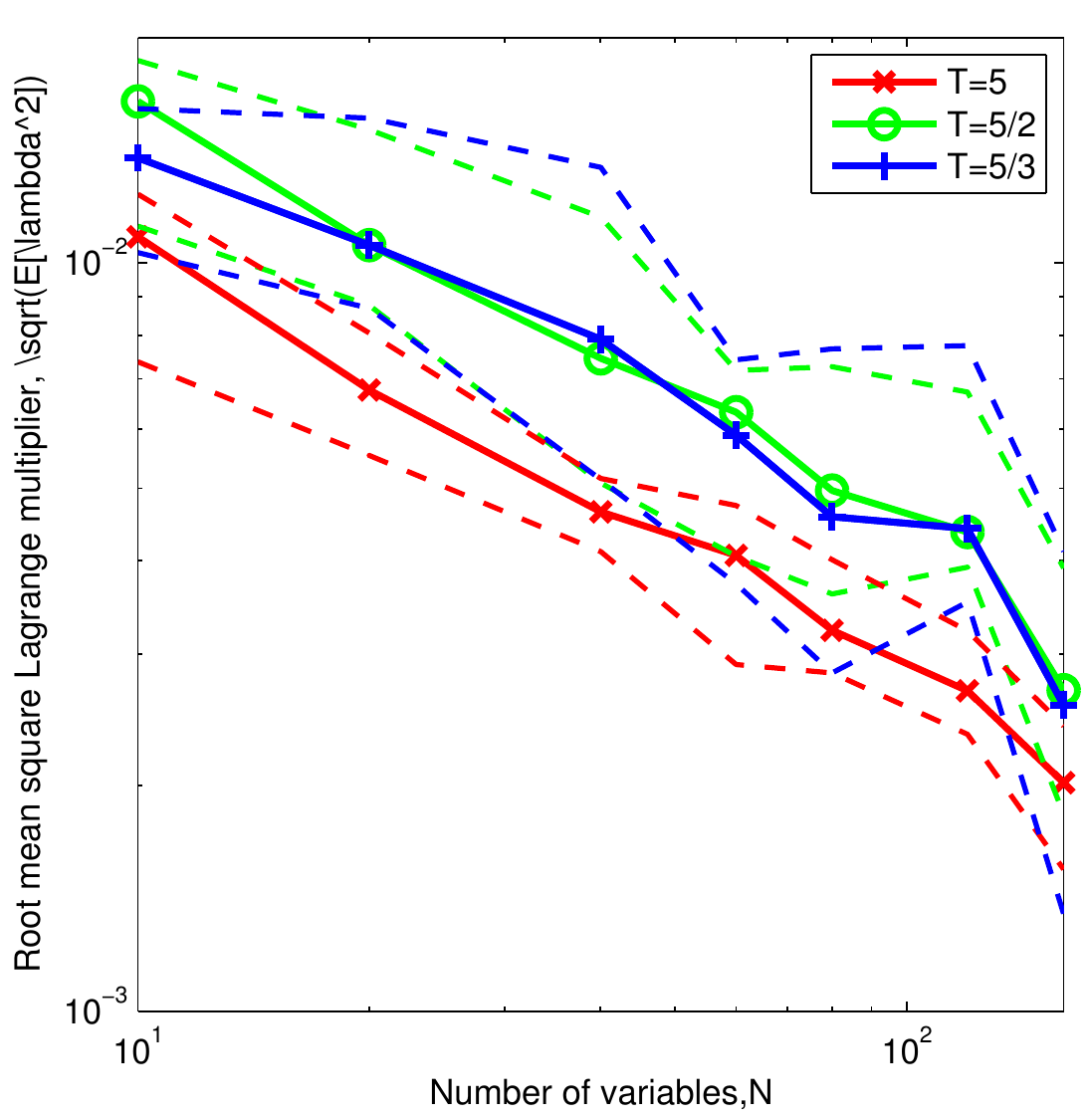}\\
\caption{\label{PottsFigure} Errors and Lagrange multipliers are shown for the 3-states Potts model on 3-regular random graphs.
    (left) The diagonal constraint
  regime yields a significant improvement in maximum absolute
  deviation of $p(x_i)$ over the raw approximation on graphs with
  $N=40$.  (right) The Bethe approximation improves as $N\rightarrow
  \infty$ at all temperatures (due to the disappearance of many short
  loops). The Lagrange multipliers enforcing diagonal consistency
  typically decrease (in mean squared value) as system size
  increase. 
  Quartiles are based on at least 20 samples per system size.}
\end{figure}
Next we consider random 3-regular graphs $G=\{V,E\}$ of $N=|V|$ variables, where each variable is allowed 3 states: $x_i\in\{0,1,2\}$. The problems are defined by the probability
\begin{equation}
  p(x) \propto \prod_{(ij) \in E} \exp\left(J_{ij}\delta_{x_i,x_j}/T\right) \prod_{i \in V}
  \exp\left(4\delta_{x_i,0}/T\right)
\end{equation}
where couplings are i.i.d.\ random variables $J_{ij}\in\{-1,1\}$.
An example of a corresponding factor graph is shown in Figure \ref{fig:alarmnet} (lower panel).
This is a disordered model, but like the fully connected Ising model, 
it is solved at leading in order in $N$ by the Bethe approximation, 
with finite size effects strongest at intermediate temperatures. At
high temperature, the probability is relatively flat and disperse,
whereas at low temperatures there is a single dominating mode concentrated about the value
$\mathbf{0}=\mathrm{argmax}_x\; p(x)$. 
 
We consider a diagonal constrained Bethe approximation: a set of $4$ non-redundant statistic pairs are constrained per variable. 

Figure \ref{PottsFigure} demonstrates that in graphs of size $N=40$,
the MAD on $p(x_i)$ is significantly improved with the addition of diagonal constraints. Furthermore, we see that as the Bethe approximation becomes accurate (high and low temperature, or large $N$) it becomes easier to enforce the constraints, as indicated by smaller values for the Lagrange multipliers.

\subsection{The alarm network}
\label{sec:alarmNet}
\begin{figure}
  \includegraphics[width=0.49\linewidth]{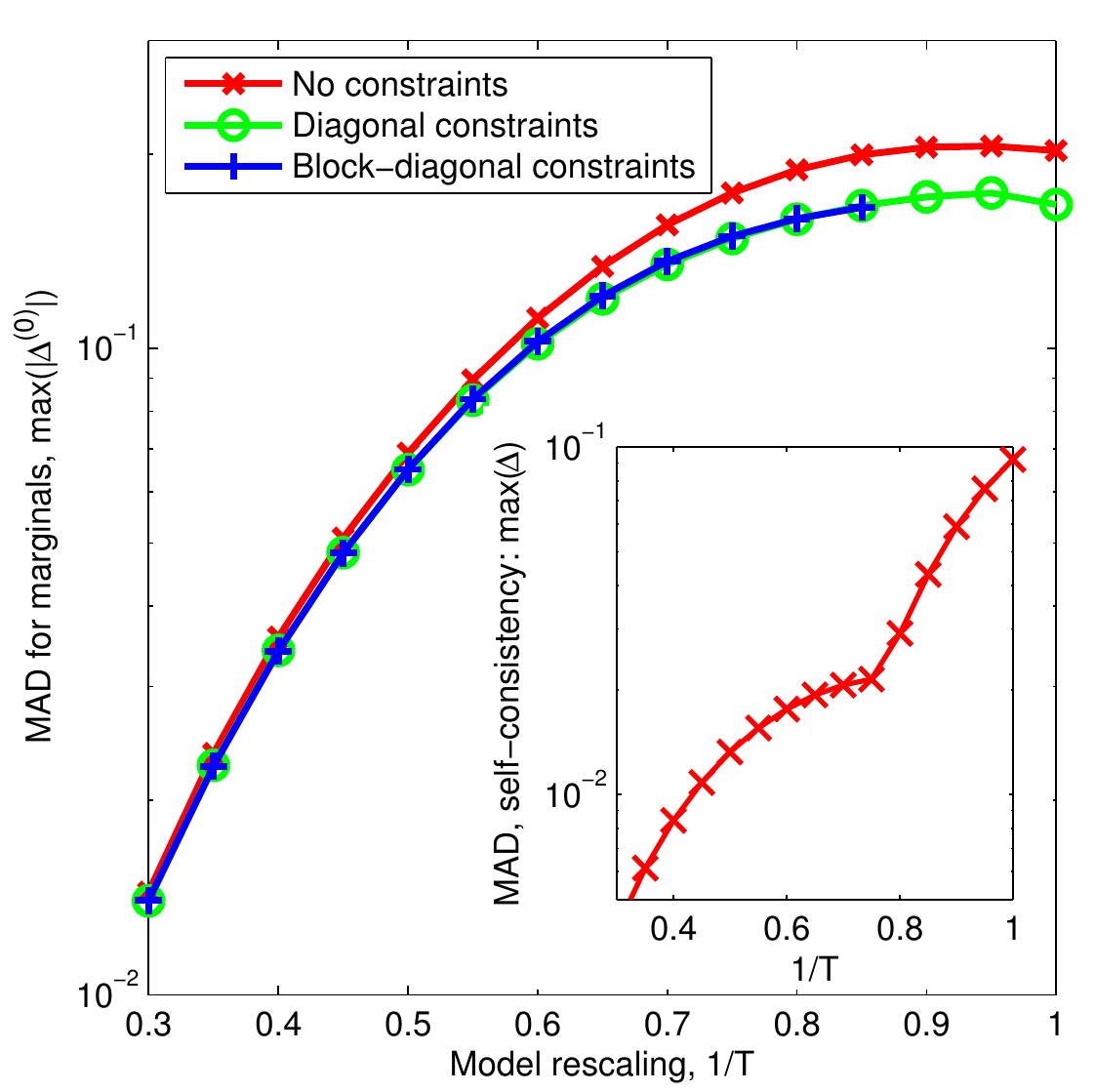}
  \includegraphics[width=0.49\linewidth]{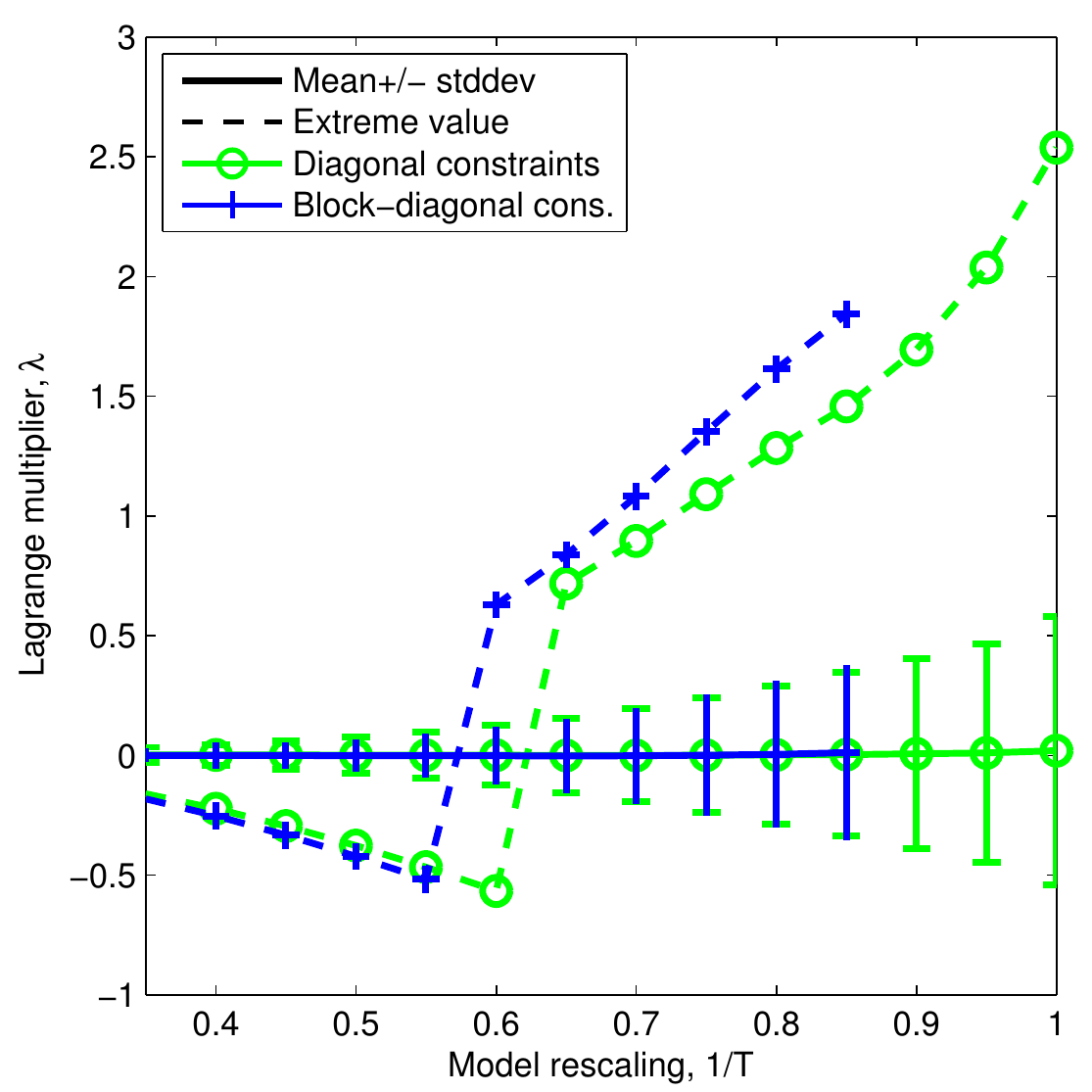}
\caption{\label{alarmNetFigure} Results for the alarm network. (left) The diagonal and block-diagonal constraint regimes have indistinguishable performance
  in the MAD for $p(x_i)$, and improve modestly on the
  unconstrained problem. (inset) At intermediate $T$ there is a qualitative change in the self-consistency error, as some variable subsets become strongly polarized. (right) The values for $\lambda$ are in the
  most part small, with extreme values diverging as $T$ approaches 1;
  a mixture of positive and negative
  Lagrange multipliers are required to enforce constraints.}
\end{figure}
The alarm net is a pedagogical example of a graphical model that has been studied in the context of loop correction
algorithms and is available in libDAI
repository~[\cite{Mooij:loopcorrected2,Mooij_libDAI_10}]. It has 37 variables, that each takes either 2, 3 or 4 states (a mixture of Ising and Potts spins). Variables have biases, and participate in 2, 3 and 4 point interactions as shown in Figure~\ref{fig:alarmnet} (upper graph). The model
involves a mixture of factor types and variables. The Bethe
approximation performs relatively poorly, due to short loops. We can
again add a temperature and consider solutions up to the full scale
$T=1$ that defines the model.

Two constraint regimes were applied: in the first pure-diagonal
constraints were applied: $\{(i,y),(i,y) : \forall i,y\}$, and in the
second all block-diagonal constraints were applied $\{(i,y_1),(i,y_2)
: \forall i,y_1,y_2\}$. The ability of both methods to improve local
statistics were comparable and modest; for both the local diagonal
and off-diagonal statistics as shown in Figure \ref{alarmNetFigure}. 
Other schemes such as tree-EP and LCBP show improvements of one or two orders of magnitude on this problem~[\cite{Mooij:loopcorrected2,Opper:IEP}]. However, the Bethe approximation and linear response estimates are known to be poor on this model - so it is not a surprise to see modest gains for our scheme. 

\section{Discussion}
\label{sec:discussion}

The aim of this paper is to show that for a range of models and variational methods we can have improved accuracy in marginal estimates by inclusion of a set of self-consistent constraints, and to propose this as a general mechanism by which to improve approximations with inconsistency between marginalization (first derivative) and linear response (second derivative) approximations. 

We have expanded upon results for rather simple approximations, in
easy to understand model frameworks: the marginals are improved in all
cases, and in many scenarios by amounts measurable in orders of
magnitude. The combination of a weak approximation method with constraints may
result in there being no solutions and in marginal cases in a slow
convergence of our proposed algorithm. On the other hand, if we are 
looking for a mechanism by which to leverage a good approximation this approach seems appropriate.

An interesting direction for investigation would be understanding
whether partially reducing the violation, rather than eliminating it,
might yield comparable results within a more robust algorithmic
framework. There are two possibilities: (1) softening the constraint, introducing a penalty term that increases with $|\Delta|$ rather than requiring strict compliance; (2) expanding the range of solution existence by bounding $\lambda$. This would presumably maintain a large part of the method advantage, removing some pathologies.
Further work could also relate to the choice of constraints,
especially for non-Ising models where the number of statistics
available to constrain is large. Two directions were proposed in Appendix \ref{app:constraintSelection}.

Some powerful approximation methods are not variational and so the insight we provide cannot be leveraged. In other cases there is no inconsistency to exploit - this is true of adaptive-TAP and some moment matching methods. These have variational free energy frameworks, but by the Gaussian approximations therein used, there is a consistency of pairwise approximations, making redundant the constraints we suggest. This is in line with our thinking that good approximations should not violate these constraints.

We believe our method also sheds light on, or is inclusive of, previous attempts to leverage linear response. Consider the region-based free energy built on the set of regions indicated by Figure \ref{fig:regions}(d). Every outer region is a hub surrounded by leaves, and if we add constraints over the leaves within each region, a linear expansion in the Lagrange multipliers agrees with the linear expansion obtained in the Montanari-Rizzo scheme~[\cite{Montanari:CLC}]; however, outside of the linearized regime there are important differences. 

There are three reasons one might not want to improve a variational approximation by adding self-consistency constraints: 1) The cost of the method which is dominated by a linear response evaluation (a cost that can be as large as $O(N^2)$ using susceptibility propagation), might be prohibitive in some applications. 2) Introducing such constraints may prevent the existence of any solutions (the method may reveal the uncomfortable truth that the approximation used is quite bad). 3) Even where a fixed point exists, it may be slow to reach it using a local iterative scheme for $\lambda$ such as (\ref{eq:iteration}).

\section{Conclusion}
\label{sec:conclusion}

We have demonstrated that adding covariance constraints, which make the linear response and marginalization estimates consistent, improves the performance of the Bethe approximation for a variety of simple model types. 
We have argued this is true more generally of variational frameworks, and have provided an algorithmic framework for mean-field and region-based frameworks, generalizing previous results.
The regimes of adding all possible constraints (on-and-off diagonal) and adding constraints only over single variable covariances (diagonal) were examined. The former tends to lead to better results, whilst the latter is simpler to implement and can yield solutions across a broader range of models. 

The usefulness of this paper is not in the specific algorithm developed but in the principle of statistical consistency. We hope it might be extended to other variational frameworks in which inconsistencies exist between first and second derivative estimates. We have presented and tested an algorithmic framework, the framework can solve the constrained free energy where solutions exist, 
but a rather expensive (annealed in $T$) procedure was exploited to obtain our results. 
Further work is required to make this reliable and competitive with state of the art in marginal estimation.

\acks{The authors thank Alejandro Lage-Castellanos for comments and Muneki Yasuda for sharing insight on the I-SUSP algorithm.}


\newpage

\appendix

\section{Exact expressions for the fully connected Ising model}
\label{app:FCIsing}
 It is straightforward to calculate the exact marginal distributions for (\ref{eq:fullConn}), for any pair of variables
%\begin{equation}
%  p(x_i) \propto \sum_{y=0}^{n-1} {n-1 \choose y}\exp\left[ \frac{(h + 2 y - n + 1)^2}{(2 T)} + \frac{h + 2 y - n - 1}{T} x_i\right] 
%\end{equation}
%and 
\begin{equation}
  p_T(x_i,x_j) \propto \sum_{n=0}^{N-2} {N-2 \choose n}\!\exp\left[ \frac{(h + [N - 2 - 2 n])^2}{2 T} \!+\! \frac{1}{T} x_i x_j \!+\! \frac{h + [N - 2 - 2 n]}{T} (x_i\! +\! x_j) \right]\;.
\end{equation}
By contrast in the Bethe approximation
\begin{equation}
  q_{ij}(x_i,x_j) \propto \exp\left[\frac{1}{T} x_i x_j + \frac{h_{\rightarrow}}{T}(x_i + x_j)\right] \label{eq:qpairBethe}
\end{equation}
where the log-ratio messages $h_{\rightarrow}$ are defined as the solution to
\begin{equation}
  h_{\rightarrow} = h + (N-1) T \mathrm{atanh}\left[\tanh(1/T)\tanh(h_\rightarrow/T)\right]\;; \qquad h_{\rightarrow} = \frac{1}{2}\log\left(\frac{\mu_{i \rightarrow (ij)}(1)}{\mu_{i \rightarrow (ij)}(-1)}\right)\;,
\end{equation}
Since we will place constraints on $\V_q(x_i,x_j)$ a convenient parameterization for the pair probability in our approximation will be
\begin{equation}
  q_{ij}(x_i,x_j) = \frac{(1 + M_i x_i)(1+M_j x_j) + C_{ij}x_i x_j}{4} \label{eq:CCbasis}
\end{equation}
We can restrict attention to symmetric solutions $M_i=M$ and $C_{ij}=C$, similarly at most two distinct Lagrange multiplier values need be considered: $\lambda_{i,i}=\lambda_0$, when diagonal constraints are applied, and $\lambda_{i,j}=\lambda_1$ when off-diagonal constraints are applied. Minimizing the variational free energy (\ref{eq:Lagrangian2}) with respect to $M$ and $C$ leads to the following pair of equations
\begin{eqnarray}
  0 \!&=&\! - [(n-1) M/T\!  +\! \lambda_0 M] + \mathrm{atanh}(M)\! +\! (N\!-\!1) \sum_{x_1,x_2}\frac{x_1}{2}q_{j}(x_2)\log\left(\frac{q_{ij}(x_1,x_2)}{q_{i}(x_1)}\right) \;. \label{eq:appSaddle1} \\
  0 \!&=&\!  - [1/T - \lambda_{1}] + \sum_{x_1,x_2}\frac{x_1 x_2}{4}\log q_{ij}(x_1,x_2) \;.\label{eq:appSaddle2}
\end{eqnarray}
If we solve this pair of equations for $\lambda=0$, we find an alternative representation of (\ref{eq:qpairBethe}).
Explicit expressions for the covariance matrix approximation $\chi$ can be derived~[\cite{Raymond:MFM,Raymond:Correcting}], these can be concisely expressed in the inverse
\begin{equation}
  [\chi^{-1}]_{i,j} = \left\lbrace\begin{array}{ll} \frac{1}{1-M^2}\left[1 + (n-1)\frac{C^2}{(1-M^2)^2 - C^2} \right] - \lambda_0 & \hbox{if }\; i=j \\ - 1/T + \sum_{x_1,x_2}\left[\frac{x_1x_2}{4}\log q_{ij}(x_1,x_2)\right] - \frac{C}{(1-M^2)^2 - C^2} & \hbox{otherwise.}
\end{array}\right.
\end{equation}
Abbreviating $\chi^{-1}_{i,i}=a$ and $\chi^{-1}_{i,j (j\neq i)}=b$ we can write the components of the covariance matrix approximation
\begin{equation}
  \chi_{i,j} = \frac{1}{(a-b)(a+(N-1)b)}\left\lbrace\begin{array}{ll} a + (N-2) b & \hbox{if }\; i=j \\ - b & \hbox{otherwise.} \end{array}\right.
\end{equation}
Thus we have up to four parameters $\{M,C,\lambda_0,\lambda_1\}$, depending which constraints are applied. We solve (\ref{eq:appSaddle1}) and (\ref{eq:appSaddle2}) in combination with either, or both, ($1-M^2=\chi_{i,i}$) and ($C_{ij}=\chi_{i,j}$). These are non-linear equations in $M$, $C$ and $\lambda_0$; we resort to local search to find solutions; solutions are possible by expansions at small or large $T$, or for large $N$. For the solution to be valid, we also check that the Hessian is positive definite, and that $p(x_i,x_j) \geq 0$.

Solutions for the equations are trivial for large or small $T$, and approach the solutions of the unconstrained approximation. By slowly varying $T$ we can discover the solutions that evolve continuously from these two fixed points. We did not find any solutions appearing discontinuously, nor did we find any evidence for symmetry breaking (i.e. solutions in which the magnetizations, or correlations, differed by label index).  

It is also straightforward to apply the same methodology for the mean-field approximation: there is one saddle-point equation. i.e.\ (\ref{eq:appSaddle1}) with $C=0$; and up to one type of covariance constraint ($\chi_{i,i}=1-M^2$), where the inverse covariance matrix elements become $[\chi^{-1}]_{i,j}=-1/T$, and $[\chi^{-1}]_{i,i}=1/(1-M^2)-\lambda_0$.
 
\section{Algorithm for finding optimized Lagrangian parameters, and the linear responses}
\label{app:Algorithms}
Our approach to find solutions depends on their existence, this is discussed generically before we consider the impact of constraint selection strategies (and constraint representation) on solution existence and algorithmic stablity.  
We then show that an exact method for minimizing the Lagrangian exists for some fixed values of the Lagrange parameters $\lambda$. The CLBP and CLSP algorithms are then presented in pseudocode. Finally, we discuss heuristics for the calculation of $\lambda$.

\subsection{Existence}
\label{app:uniqueness}

Certain models are known to be solved at leading order in some control parameter by region-based, Bethe or mean-field approximation. Examples are ferromagnetic random graphs or fully connected models in the limit of large number of variables ($N$), arbitrary models in the weakly interacting limit (also called high temperature $T$), finite models in the limit of strong biases on the variables ($h$). In these models, we find, of course, that the deviation between the linear response and marginal estimates are small in the corresponding control parameter. Expanding in this parameter (e.g. $1/T$, $1/N$) it is straightforward to show the existence of solutions, and quantify the effectiveness of the constraints~[\cite{Raymond:Correcting}]. However, even outside the regime where expansions about the Bethe approximation are appropriate, we find solutions: this includes models where the covariance matrix has divergent terms, due to a mean-field phase transition~[\cite{Raymond:MFM}]. 

However, in experiments, we present it is shown that variation of the temperature (which controls the smoothness of the probability) can lead to models with no solutions, for some constraint sets. In the case of both on and off-diagonal constraints in Ising models defined on some graph $G=\{V,E\}$, with $|V|$ variables and $|E|$ edges, we have $|E|+|V|$ variational parameters but also $|E|+|V|$ constraints - solving a system of non-linear equations where the number of parameters matches the number of constraints seems optimistic; and perhaps it should not be surprising that as the Bethe approximation breaks down solutions fail to exist; applying all constraints is certainly a marginal case unless some are redundant. If we move to alphabets with more than $2$ states per variable the number of pair-statistics for on-and-off diagonal constraints exceeds the number of variables (with no obvious redundancy); certainly such a system can have no solutions. We must think carefully on which constraints to implement, and this is discussed in Appendix \ref{app:constraintSelection}.

In the experiments of this paper we have introduced for all models the control parameter $T$, with all models being well approximated by Bethe (and Mean-field) in the limit $T\rightarrow \infty$, and some models also being well approximated with $T\rightarrow 0$. To extend the regime in which our algorithms are convergent, an annealing approach was taken - slowly decreasing (increasing) $T$ and following a solution which evolved continuously in the marginals and Lagrange multipliers. In many cases, solutions were found to reach a critical point $T^*$ beyond which they could not be continuously evolved to new solutions - and at these points invariably we found no new solutions emerging discontinuously by simply iterating our procedures. By undertaking annealing, it would in principle be possible to miss some solution that emerges discontinuously; the example of Section \ref{sec:FC} shows this is possible (there is coexistence of two solutions at low temperature for the case of off-diagonal constraints only), but in the other examples of this paper we found no evidence for this, and we expect in practical applications and for good choices of the constraints coexistence will be absent. 

A common feature in solution discontinuity during annealing is the divergence of some Lagrange multiplier(s); this indicates that the failure was due to inviability of solutions - rather than a breakdown of algorithmic dynamics. Unlike the Bethe approximation, which can be forced to converge to a local minimum at any $T$ (e.g. replacing LBP by a convex-concave procedure). We speculate that the constrained free energy has no solutions in strongly coupled regimes and that where practical solutions do exist the cavity based algorithms will be sufficient if combined with appropriate damping and/or annealing.

\subsection{Selection of constraints for best solutions}
\label{app:constraintSelection}

Independent of the issues of solution existence and algorithmic convergence, we consider in this section reasonable choices for constraints.

Since we are applying constraints to discrete models it is clear that the apriori constraints selection should be independent of the labeling convention. Furthermore, we note that we made the rather arbitrary choice in (\ref{eq:prob}) of perturbations in the set of statistics $\{\delta_{x_i,y} : y = 1\ldots Y\}$ , $Y$ being the number of states. Whilst this is invariant under relabeling, and spans all possible perturbation directions (and one redundant direction $\sum_y \delta_{x_i,y}$), we might get different covariances according to our choice. It would seem sensible to choose our constraint regime so that any set of statistics that spans the set of perturbations yields the same result. To achieve this we must pick a complete basis $\{\phi\}$, and apply constraints on all-covariance pairs. 

The consideration of bases motivates the regimes we have explored: diagonal, off-diagonal, on-and-off diagonal. These schemes are basis-independent. If the condition $C_{(i,\cdot),(j,\cdot)}=\chi_{(i,\cdot),(j,\cdot)}$ is met in one basis for all elements, then a change between two (orthonormal) basis is a rotation, and the identities remain intact.

We found that off-diagonal typically performed worse than on-diagonal, as well as being computationally more challenging.  In the final experiment of Section~\ref{sec:alarmNet} we also tried one regime that was basis dependent, the so-called {\em pure-diagonal} regime. Despite breaking the paradigm of basis independence, it performed just as well as the {\em block-diagonal} regime that is basis independent, so this indicates that basis independence is not an essential feature in constraint selection. 

Topological distinctions amongst constraints can also be made. In a region-based approximation, we can distinguish the ``2-core'' from the rest of the graph. The 2-core is the set of variables that are found by recursively removing variables that are involved in at most one ``outer region'' (generalized interaction). We advise constraining variables on the 2-core only. The reason for this is that the set of variables that are not in the 2-core form a forest - a set of disconnected trees for which the approximation is conditionally exact even without constraints. If we have the core correct, then other constraints will be redundant. It is only in the case of the alarm net (Figure \ref{fig:alarmnet}, Section \ref{sec:alarmNet}) that the 2-core is distinguished from the full graph and we make this approximation. Other topological distinctions could be applied, distinguishing constraints by the distance between variables might be one possibility (e.g. Figure \ref{fig:regions}(b) there are nearest, and next nearest neighbors); however, in the Bethe approximation (all experiments we present) only nearest neighbors might be constrained so the example is mute. 

Beyond these considerations, we did not attempt to further restrict the set of constraints in a model specific manner, but several options might be worth exploring, in particular: (1) Consider the covariance matrix restricted to a single variable ``block'', $\chi_{(i,\cdot),(i,\cdot)}$. For a given approximation (e.g. Bethe) there is a unique orthonormal basis $\{\phi_i\}$ that diagonalizes the covariance matrix. This would provide a natural choice for the statistic basis. The eigenvalue-vector pairs determine which direction is most susceptibility to a change in parameters, and so (loosely speaking) most sensitive to approximation errors. We might, therefore, rank constraints by this eigenvalues for inclusion. (2) If we can solve the problem to obtain $q_i^*$, then we can consider defining only one perturbation statistic per variable as $\phi(x_i)\propto \log(Y q^*(x_i))$. Since we know approximations tend to be overconfident, it would seem a natural choice to constrain in line with the belief, rather than wasting resources on unimportant directions in parameter space.

\subsection{Assignment of constraints to regions, and basis selection, for best convergence}
\label{app:constraintAssignment}

For our constraint regimes (diagonal, on-diagonal and on-and-off diagonal),  the allocation of constraints to specific regions, the choice of statistical basis, the initial condition of the algorithm and the damping, leave the algorithmic fixed points unchanged. Owing to our slow annealing approach (increasing or decreasing $T$) our results were not very sensitive to the other implementation details. However, these choices do impact convergence significantly. 

In terms of the statistics basis, choosing a non-redundant orthonormal set was found to lead to fastest convergence. Orthonormality ensures that, at leading order, the Lagrange multipliers are independent of one another, which is the criterium that guarantees the success of our method. A redundant statistic is $\phi(x_i)=\sum_{y=0}^{Y-1} \delta_{x_i,y}$, where $x_i$ has $Y$ states, and so only $Y-1$ orthonormal statistics need be implement. A nice approach to the selection of an orthonormal basis is discussed in Yasuda et al.~[\cite{Yasuda:GISP}], for Ising spins, a unique choice per variable is $x_i$.

We allocated constraints in a naive manner, assigning each constraint to the first available ``outer region''. For diagonal constraints an approach where constraints are allocated to single-variable regions was proposed by Yasuda et al., and this allows some simplifications as well as implying a unique allocation~[\cite{Yasuda:GISP}]. Approaches that do not allocate constraints to regions, such as iteration by linear expansion of the covariance matrix have been attempted (\ref{eq:LinExpChi}), but those were slower to converge. 

As a general rule, we wish to group Lagrange multipliers together as much as possible. In this way, correlations amongst the Lagrange multipliers are accounted for locally. A potentially better approach than our fixed assignment suggestion might be to update all Lagrange within some outer region. Where a constraint could be allocated to more than one region, the results could be averaged, or regions could be selected differently on each cycle to avoid conflicting assignments. 

A damping factor that begins as $1$ and decreases during the annealing procedure (as required to enable convergence) is applied in the experiments. In simulations without an annealing scheme, a rate $\lesssim 0.5$ could be essential to prevent oscillations that arise from correlated updating of Lagrange multipliers. In particular, off-diagonal values $\lambda_{ij}$ and diagonal values $\lambda_{ii}$ can be anti-correlated. If they are not updated on the same region undamped iteration can lead to slow convergence or oscillations.

Initial conditions are chosen as $\lambda=0$; so that the first iteration is equivalent to an unconstrained approximation. One advantage of the annealing approach (where we first solve the trivial model $T=0$ or $+\infty$, and then increase or decrease $T$, initializing by the current solution), is that we do not require the convergence of the unconstrained approximation to find a solution in the constrained regime.

\subsection{Convex Concave procedure}
\label{eq:CCp}
We prove that it is possible to minimize (\ref{Lagrangian}) in $q$ by a convex-concave decomposition.
In~[\cite{Heskes:AICO}] by introducing auxiliary parameters $q'$ over the inner regions, a pair of convex optimization procedures were developed that led to minima of the region based free energy. With the addition of our constraints, we need only make a minor modification to their argument. Consider an auxiliary free energy $F(q,q')$, where $q'$ are additional parameters in one to one correspondence with the original variational parameters. Heskes et al. outline three requirements of this free energy
\begin{itemize}
  \item[1] Convexity of $F(q,q')$ with respect to $q$.
  \item[2] $F(q,q'=q) = F(q)$
  \item[3] $F(q,q') \geq F(q)$
\end{itemize}
that guarantee that the iterative procedure
\begin{equation}
  q'_{t+1} = \mathrm{argmin}_{q} F(q,q'_t) \label{itscheme}
\end{equation}
converges to values that are minimizing arguments of $F(q)$.

Since the violation term (\ref{eq:Delta}) is a quadratic function of the variational parameters $q(x_\alpha)$, it can be expressed by a constant $A_0$, a vector of linear coefficients $A_1$ and quadratic coefficients $A_2$ that define a symmetric matrix. $A_2$ can be decomposed as a positive definite and negative semi-definite part ($A_2^+$ and $A_2^-$). It is then easy to define a function in vector notation
\begin{equation}
\Delta^+(q,q',q^*_{(\cdot)}) = A_0(q^*_{(\cdot)}) + q^T A_1 +  (q')^T A_2 q' + (q-q')^T A_2^+ (q-q') \label{eq:Deltaplus}
\end{equation} 
To minimize our region-based free energy with constraints (\ref{Lagrangian}) we substitute (\ref{eq:Deltaplus}) for (\ref{eq:Delta}) and follow the procedure in~[\cite{Heskes:AICO}] for the remaining terms. The free energy meets the criteria for convergence under the iterative scheme (\ref{itscheme}).

The double loop procedure, despite not changing the asymptotic complexity of message passing $O(N)$ is often considered impractical, but, in the framework we are proposing, the largest cost is the evaluation of the linear response, implementing a double loop procedure may be sensible. In practice we have used the variation on loopy belief propagation, rather than the double-loop procedure, to extremize the Lagrangian. Despite the absence of theoretical guarantees, the 3-fold scheme proves to be reliable: solving for variational parameters (by CLBP); solving for the linear response (by CLSP); and update of the Lagrange multipliers.

\subsection{Pseudocode for determining the fixed point of q at fixed $\alpha$}
\label{app:pseudocode}

We determine $q$ by the constrained loopy belief propagation (CLBP) algorithm, described in Algorithm~\ref{HAKalg}, which is a modified form of the loopy belief propagation algorithm~[\cite{Heskes:AICO}], that uses the convex-concave procedure with guaranteed convergence.

In Algorithm~\ref{HAKalg} we consider an auxiliary region interaction including both interactions and the Lagrange multiplier terms
\begin{equation}
  \psi_\alpha(x_\alpha;\boldsymbol{\lambda}_\alpha,q_\alpha) = \psi_\alpha(x_\alpha)
  \prod_{[(i,s_1),(j,s_2)] \in \omega_\alpha} \exp\left\lbrace - \lambda_{(i_1,y_1),(i_2,y_2)} {\hat \phi}_{(i_1,y_1),(i_2,y_2)}(x_\alpha; q_\alpha)\right\rbrace \;,
\end{equation}
and
\begin{multline}
  {\hat \phi}_{(i_1,y_1),(i_2,y_2)}(x_\alpha; q_\alpha) = \left(\delta_{x_{i_1},y_1} - \E_{q_\alpha}(\delta_{x_{i_1},y_1}) \right) \left(\delta_{x_{i_2},y_2} - \E_{q_\alpha}(\delta_{x_{i_2},y_2})\right) \\
+  \E_{q_\alpha}(\delta_{x_{i_2},y_2})\E_{q_\alpha}(\delta_{x_{i_1},y_1}) \;.
\end{multline}
Simplified expressions for the Ising model are presented in the main text. Terms here have an interpretation compatible with~[\cite{Heskes:AICO}], with $U$ are the set of {\em outer regions}, and $V$ are the set of intersection regions (see Figure \ref{fig:regions}). In the case of a Bethe approximation $U$ are the set of edges, and $V$ are the set of variables.

\begin{algorithm}[H]
\caption{\label{HAKalg} $\lambda$ compatible Heskes-Albers-Kappen algorithm}
\begin{algorithmic}
    \WHILE{$\neg$ converged}
    \STATE \FORALL{$\beta \in V$}
    \STATE \FORALL{$\alpha \in U,\alpha \supset \beta$}    
    \STATE \begin{equation} 
      q_\alpha(x_\beta) = \sum_{x_{\alpha\setminus\beta}} q_\alpha(x_\alpha) \label{eq1}
    \end{equation}
    \STATE \begin{equation} 
      \mu_{\alpha\rightarrow\beta}(x_\beta) = \frac{q_\alpha(x_\beta)}{\mu_{\beta\rightarrow\alpha}(x_\beta)} \label{eq2}
    \end{equation}
    \ENDFOR
    
    \STATE \begin{equation}
      q^{(num)}_\beta(x_\beta) = \prod_{\alpha \supset \beta}\mu_{\alpha \rightarrow \beta}(x_\beta) \label{eq3} 
    \end{equation}
    \STATE \begin{equation}
      q_\beta(x_\beta) = \frac{q^{(num)}_\beta(x_\beta)}{\sum_{x_\beta'} q^{(num)}_\beta(x_\beta')} \label{eq4}
    \end{equation}
    
    \STATE \FORALL{$\alpha \in U,\alpha \supset \beta$}

    \STATE \begin{equation}
      \mu_{\beta \rightarrow \alpha}(x_\beta) = \frac{q_\beta(x_\beta)}{\mu_{\alpha \rightarrow \beta}(x_\beta)} \label{eq5}
    \end{equation}

    \STATE \WHILE{$\neg$ converged (solve self-consistently)}
    \STATE \begin{equation}
      q^{(num)}_\alpha(x_\alpha) = \psi_\alpha (x_\alpha; \lambda, q_{\alpha}) \prod_{\beta \subset \alpha} \mu_{\beta \rightarrow \alpha}(x_\beta)\label{eq6}
    \end{equation}
    \STATE \begin{equation}
      q_\alpha(x_\alpha) = \frac{q^{(num)}_\alpha(x_\alpha)}{\sum_{x_\alpha'} q^{(num)}_\alpha(x_\alpha')}\label{eq7}
    \end{equation}
    \ENDWHILE

    \ENDFOR
    \ENDFOR
    \ENDWHILE
\end{algorithmic}
\end{algorithm}

\subsection{Linear response about the fixed point, fixed $\lambda$}
\label{app:linearresponse}

By linearizing the CLBP method, we obtain a constrained loopy susceptibility propagation (CLSP) method. Qualitatively similar is the I-SUSP algorithm~[\cite{Yasuda:SPDC}], which applies only to the case of diagonal constraints and the Bethe approximation. 

The implementation we use follows Algorithm~\ref{HAKalg}, but with the following replacements for each equation (in order). 
\begin{eqnarray}
  \delta {\hat q}_{\alpha}(x_\beta) &=& \sum_{x_{\alpha\setminus\beta}} \frac{q_{\alpha}(x_\alpha)}{q_{\beta}(x_\beta)} \delta {\hat q}_{\alpha}(x_\alpha) \label{eq1d}\\
  \delta {\hat \mu}_{\alpha\rightarrow\beta}(x_\beta) &=& \delta {\hat q}_{\alpha}(x_\beta) - \delta {\hat \mu}_{\beta\rightarrow\alpha}(x_\beta) \label{eq2d}\\
  \delta {\hat q}^{(num)}_\beta(x_\beta) &=& \sum_{\alpha \supset \beta} \delta {\hat \mu}_{\alpha \rightarrow \beta}(x_\beta)  \label{eq3d}\\
  \delta {\hat q}_\beta(x_\beta) &=& \delta {\hat q}^{(num)}_\beta(x_\beta) - \sum_{x_\beta} q_\beta(x_\beta) \delta {\hat q}^{(num)}_\beta(x_\beta) \label{eq4d}\\
  \delta{\hat \mu}_{\beta \rightarrow \alpha}(x_\beta) &=& \delta {\hat q}_\beta(x_\beta)  - \delta {\hat \mu}_{\alpha \rightarrow \beta}(x_\beta) \label{eq5d}\\
  \delta {\hat q}^{(num)}_\alpha(x_\alpha) &=& \frac{\delta_{x_i,y} {\mathbb I}(i \in \alpha)}{k_i} + \sum_{\beta \subset \alpha} \delta {\hat \mu}_{\beta \rightarrow \alpha}(x_\beta) \nonumber \\ &+& \sum_{x_\alpha'}q_\alpha(x_\alpha') \delta {\hat q}_\alpha(x_\alpha') \frac{\partial \log \psi(x_\alpha ;\boldsymbol{\lambda}_\alpha,q_\alpha)}{\partial q_\alpha(x_\alpha')}\label{eq7d}\\
  \delta {\hat q}_\alpha(x_\alpha) &=& \delta {\hat q}^{(num)}_\alpha(x_\alpha) - \sum_{x_\alpha'} q_\alpha(x_\alpha') \delta {\hat q}^{(num)}_\alpha(x_\alpha') \label{eq8d}\;,
\end{eqnarray}
where ${\mathbb I}(i \in \alpha)$ is an indicator function evaluating to one if $i$ is contained in $\alpha$, and $k_i$ is the number of outer regions containing $i$. From the converged quantities we can identify
\begin{equation}
  q^*_{\alpha,(i,y)}(x_\alpha) = q_\alpha(x_\alpha) \delta {\hat q}_\alpha(x_\alpha) \label{eq:dqstar}\;.
\end{equation}
Since the linear response is a linearized method, the innermost do while loop of Algorithm~\ref{HAKalg} could be replaced by a method for solving linear equations. Both have been tried with similar outcomes.

\section{Solving for $\lambda$, explicit expressions and simplification}
\label{app:solvingforlambda}

The diagonal update scheme of I-SUSP relies on a similar reasoning as that undertaken in this paper~[\cite{Yasuda:SPDC,Yasuda:GISP}]. The difference between the two approaches is that in the works by Yasuda et al. only diagonal constraints (restricted to single variable consistency relations) are considered. As such it is possible to choose in general a single variable cavity $i$, rather than a region cavity $\alpha$ (see Figure \ref{fig:cavityargument}). To consider statistics not restricted to single variables our scheme is required. The I-SUSP method has an advantage over our method in that there is a simplification of the expressions on a single variable region that allows a closed form for updating the Lagrange multipliers. One advantage of our scheme is that we may fix several Lagrange multipliers on a region simultaneously, thus accounting for correlations in their values, and perhaps reducing the number of updates to convergence. 

In practice, damping within the 3-cycle of Figure \ref{fig:3cycle} can be necessary for convergence. In practice, we replace $\lambda^{t+1} = d\lambda^{t} + (1-d) \lambda^*$, where $\lambda^*$ is the cavity approximation. As $1/T$ was increased (or decreased) at a constrant rate we increased $d$ whenever the solution failed to converge, and this modification extended the region of convergence. 

An exact expression for the correlation matrix $\chi$ as a function of the variational parameters was derived in~[\cite{Raymond:MFM,Raymond:Correcting}], the special case for homogeneous graphs is made explicit (Appendix \ref{app:FCIsing}). The inverse covariance matrix $\chi^{-1}$ is, in the basis where variational parameters are beliefs $q$, a linear function of $\lambda$. Updating a single $\lambda$ can be achieved by the Shermann-Morrison formula, which in its linearized form becomes
\begin{equation}
   \chi_{i,i}\,\delta \lambda_{i,j}\,\chi_{j,j} = (C_{i,j}^* - \chi_{i,j})\;. \label{eq:LinExpChi}
\end{equation}
It is also possible to write a similar expression blockwise (over $\boldsymbol{\lambda_\alpha}$).
This update scheme and variations were considered, but were found to require more iterations or stronger damping (for convergence) than the linearized cavity formula (\ref{eq:iteration}). 

In our approach we associate each constraint to a unique region, but some statistics such as $\V_p(x_i,x_i)$ may be approximated by different $\alpha$ for the same variational approach 
(e.g by $q_{ij}$ or $q_{ik}$ if $i$ participates in two
interactions). Though these estimates agree finally at the solution point, they disagree at intermediate stages of CLBP. Since updates for $\lambda$ may be correlated (consider Figure \ref{fig:JordanWainwrightLambda} where diagonal and off-diagonal are strongly anti-correlated), the grouping of $\lambda$ may be very important. Updating non-disjoint sets and averaging over results may be more stable than our scheme where closely correlated $\lambda$ (e.g. $\lambda_{i,i}$ and $\lambda_{i,j}$) may be updated on different regions.

%We apply this scheme in parallel to every $\alpha$ -- since some constraints can be resolved on multiple regions (e.g. $\delta \lambda_i$ can be determined differently on region $ij$ and $ik$ at intermediate stages), we can average the estimates or simply consider variation of $\lambda$ restricted to one region (each constraint is assigned to one region).

%In the general scheme we solve for $\lambda$ by linearizing equation (\ref{eq:iteration}) about the current solution. The deviation in the statistic $(i_1,s_1),(i_2,s_2)$, the right hand-side of the equation, can be determined from $q^*_{\alpha}$ and $q^*_{\alpha,(i_1,s_1)}$. The left hand-side involves derivatives of these same statistics, the case of $C$ is relatively simple at leading order in $\lambda$
%\begin{equation}
%  \frac{\partial C_{\omega}}{\partial \lambda_{\omega'}} = \frac{-1}{1+\delta_{\omega_1',\omega_2'}}\left( \sum_{q^*}[{\hat \phi}_{\omega}(x_\alpha; q^*_\alpha){\hat \phi}_{\omega'}(x_\alpha; q^*_\alpha)] -  \sum_{q^*}[{\hat \phi}_{\omega}(x_\alpha; q^*_\alpha)]\sum_{q^*}[{\hat \phi}_{\omega'}(x_\alpha; q^*_\alpha)]\right) \label{eq:}
%\end{equation}
%whereas to solve for derivative of $\chi_{\omega}$, we have the linear equation that determines $\delta q^*_{\omega_1}$ (\ref{eq:dqstar}), and can take the derivative with respect to $\lambda$. In practice we need to incorporate damping into the update rule for strongly interacting models. Without damping the domain of convergence is substantially reduced in the figures of Section \ref{sec:results}.

\vskip 0.2in
\bibliography{Bibliography}
\end{document}